\documentclass[runningheads]{llncs}

\usepackage{eccv}

\usepackage{eccvabbrv}

\usepackage{graphicx}
\usepackage{booktabs}

\usepackage{tikz}
\usepackage{comment}
\usepackage{amsmath,amssymb} 
\usepackage{color}
\usepackage{multirow}
\usepackage{colortbl}
\usepackage{tcolorbox}
\usepackage{mleftright}
\usepackage{url}

\usepackage{wrapfig}
\def\tabref#1{Table \ref{#1}}
\def\figref#1{Fig. \ref{#1}}
\def\secref#1{Section \ref{#1}}
\def\eqref#1{Eq (\ref{#1})}
\def\Bdma#1{\mbox{\boldmath{$#1$}}}
\def\ScoreUp#1{\textcolor{red}{#1}}
\def\ScoreDown#1{\textcolor{blue}{#1}}

\definecolor{deepgreen}{rgb}{ .0, .44, .0}

\usepackage[accsupp]{axessibility}  

\usepackage[pagebackref,breaklinks,colorlinks]{hyperref}

\usepackage{orcidlink}
\usepackage{caption}

\begin{document}

\title{Vision-Language Models Learn Super Images for Efficient Partially Relevant Video Retrieval} 

\titlerunning{VLMs Learn Super Images for Efficient PRVR}

\author{Taichi Nishimura\inst{1} \and Shota Nakada\inst{1} \and Masayoshi Kondo\inst{1}}

\authorrunning{T. Nishimura et al.}

\institute{LY Corporation \\
\email{\{tainishi,shota.nakada,masayoshi.kondo\}@lycorp.co.jp}}

\maketitle

\begin{abstract}
In this paper, we propose an efficient and high-performance method for partially relevant video retrieval, which aims to retrieve long videos that contain at least one moment relevant to the input text query.
The challenge lies in encoding dense frames using visual backbones. This requires models to handle the increased frames, resulting in significant computation costs for long videos.
To mitigate the costs, previous studies use lightweight visual backbones, yielding sub-optimal retrieval performance due to their limited capabilities.
However, it is undesirable to simply replace the backbones with high-performance large vision-and-language models (VLMs) due to their low efficiency.
To address this dilemma, instead of dense frames, we focus on super images, which are created by rearranging the video frames in an $N \times N$ grid layout.
This reduces the number of visual encodings to $\frac{1}{N^2}$ and mitigates the low efficiency of large VLMs.
Based on this idea, we make two contributions.
First, we explore whether VLMs generalize to super images in a zero-shot setting. To this end, we propose a method called query-attentive super image retrieval (QASIR), which attends to partial moments relevant to the input query.
The zero-shot QASIR yields two discoveries: (1) it enables VLMs to generalize to super images and (2) the grid size $N$, image resolution, and VLM size are key trade-off parameters between performance and computation costs.
Second, we introduce fine-tuning and hybrid QASIR that combines high- and low-efficiency models to strike a balance between performance and computation costs.
This reveals two findings: (1) the fine-tuning QASIR enhances VLMs to learn super images effectively, and (2) the hybrid QASIR minimizes the performance drop of large VLMs while reducing the computation costs.
\keywords{Vision-language models, Super images, Partially relevant video retrieval}
\end{abstract}

\section{Introduction}
\label{sec:intro}

\begin{figure}[t]
  \centering
  \includegraphics[width=0.78\linewidth]{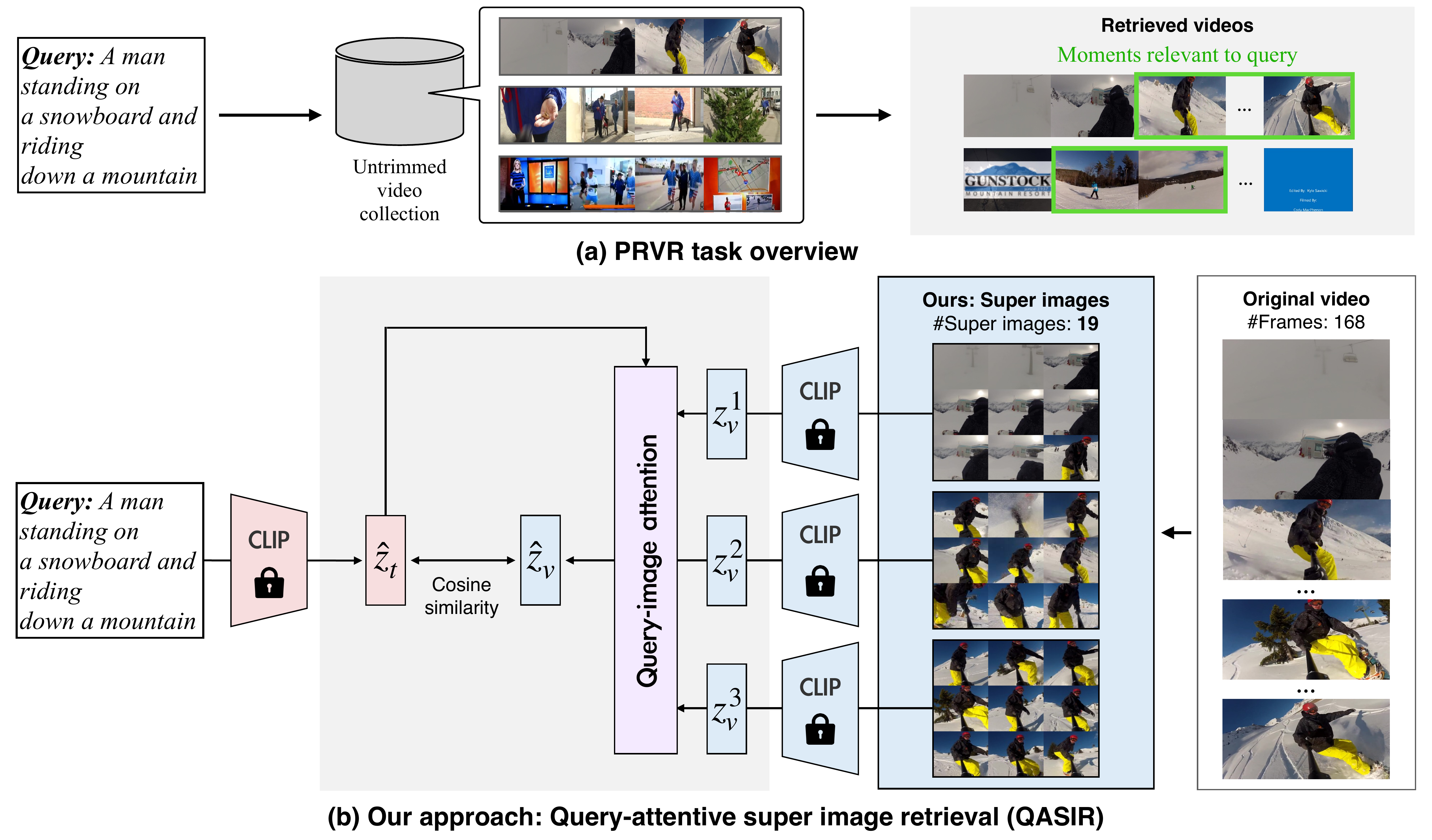}
  \caption{(a) Overview of PRVR task. (b) Key concept of our approach. Super images are created by rearranging frames in $N \times N$ grid layout, enabling the model to reduce number of encodings through visual backbones.}
  \label{fig:task_overview}
\end{figure}

With the explosive increase of videos on the web, text-to-video retrieval (T2VR), a task of searching videos relevant to an input text query, has garnered significant attention in vision and language \cite{chen2020cvpr,dong2018tmm,han2021acmmm,hu2022eccv,liu2021acmmm,song2019cvpr,wang2020tmm}.
The mainstream T2VR assumes that videos are pre-trimmed and short in duration and that the input text queries are single sentences fully relevant to the videos.
However, in practice, videos are generally untrimmed and long in duration and contain multiple moments \cite{krishna2017iccv,zhou2018aaai,zhang2023iccv,bao2021aaai}.
For this reason, the input query typically pertains to specific moments within the videos, rather than encompassing their entire content.
To fit T2VR to a real-world scenario, partially relevant video retrieval (PRVR) was proposed \cite{dong2022acmmm}, which aims at retrieving videos that contain at least one moment relevant to the query (\figref{fig:task_overview} (a)).
Previous studies proposed a method for linking a text query with dense frame representations \cite{dong2022acmmm,dong2023iccv}.

For practical use for PRVR, it is essential to achieve both high computational efficiency and retrieval performance.
From both perspectives, the challenge lies in the encoding of dense frames through visual backbones.
They require a model to encode the increased number of frames, leading to a rise in computation costs, especially for long, untrimmed videos.
To mitigate the costs, previous methods chose lightweight backbones such as ResNet152 \cite{he2016cvpr} and CLIP-B/32 \cite{radford2021icml}.
This backbone choice may yield performance issues due to its limited capabilities for learned visual representations compared with large vision-and-language models (VLMs), such as CLIP-L/14.
A recent study demonstrated that VLMs follow a scaling law \cite{kaplan2020arxiv}, wherein increased model sizes correlate with enhanced performance on downstream tasks; thus, larger VLMs are expected to improve the retrieval performance for PRVR.
However, simply replacing the backbone with such VLMs is undesirable because they suffer from low efficiency, consuming much higher GFLOPs.
For example, CLIP-L/14 \cite{fang2023cvpr} consumes 162.0 GFLOPs to encode a single image, which is considerably higher than the lightweight backbones of ResNet152 ($=$11.6 GFLOPs) and CLIP-B/32 ($=$8.8 GFLOPs).
Therefore, a dilemma arises in balancing computational efficiency and high performance when leveraging large VLMs.

One possible solution is to reduce the number of visual encodings.
For efficient video processing \cite{lei2021cvpr,lei2023acl,wang2023cvpr,Zhi_2021_ICCV,Xian_2024_WACV}, methods have proposed for sparsely sampling frames based on randomness \cite{lei2021cvpr,lei2023acl}, motion \cite{Zhi_2021_ICCV}, and frame similarity \cite{Xian_2024_WACV}, demonstrating their effectiveness in tasks such as T2VR and action recognition.
However, these methods are optimized for short videos or situations where the queries are fully relevant to the videos. Hence, they have not been thoroughly validated for PRVR. Regrettably, they show a sharp performance drop for PRVR as the number of sampling frames decreases on benchmark datasets (\secref{subsec:comparison_to_other_approaches}).
This is because a model is required to focus on local moments in the videos, and these sparse sampling methods may miss them.

In this study, we overcome these difficulties and propose an efficient and high-performance PRVR method that uses large VLMs (\figref{fig:task_overview} (b)).
Instead of using dense frames or sampling frames sparsely, we focus on super images \cite{fan2022iclr}, which are created by rearranging video frames in an $N \times N$ grid layout.
This reduces the number of visual encodings to $\frac{1}{N^2}$ and mitigates the low efficiency of large VLMs, allowing us to adopt them as a powerful image-text encoder.
Based on this idea, we address two research questions: \textit{(1) Do VLMs generalize well to super images for PRVR?} and if yes, \textit{(2) how can we achieve an efficient and high-performance method by combining them and super images?}

To answer (1), we investigate whether VLMs generalize to super images in a zero-shot setting. For this purpose, we propose a method called query-attentive super image retrieval (QASIR), which attends to partial moments relevant to the input query for PRVR. The zero-shot QASIR yields two discoveries. First, it enables VLMs to generalize to super images and achieve promising performance against fully-trained SOTA methods with conventional backbones.
Second, the grid size, image resolution, and VLM size are trade-off parameters between performance and computation costs. In other words, opting for a smaller grid size, higher resolution, and larger VLMs enhances the performance but requires increased computation costs, and vice versa.

To answer (2), we introduce a fine-tuning QASIR that incorporates a few trainable modules and a hybrid QASIR that combines high- (larger $N$ and smaller VLMs) and low-efficiency models (smaller $N$ and larger VLMs) to strike a balance between retrieval performance and computation costs. The fine-tuning experiments reveal two findings.
First, the fine-tuning QASIR enables the VLMs to learn super images effectively and achieve comparable performance to VLMs that use dense frames while reducing computation costs. Second, the hybrid QASIR minimizes the performance drop of large VLMs, further reducing the computation costs. We will release the code upon the paper acceptance.

\section{Related work}
\label{sec:related_work}

\textbf{Text-to-video retrieval (T2VR)} \cite{chen2020cvpr,dong2018tmm,han2021acmmm,hu2022eccv,liu2021acmmm,song2019cvpr,wang2020tmm,dong2021pami} aims at searching for relevant videos from an input text query. The traditional T2VR studies evaluate methods on datasets consisting of short videos and fully-relevant queries \cite{chen-dolan-2011-collecting,msrvtt,Wang_2019_ICCV,Rohrbach_2015_CVPR}. Hence, there exists a gap between the traditional T2VR and practical use because the real videos are often untrimmed and lengthy, while the queries are partially relevant.
\\\textbf{Partially relevant video retrieval (PRVR)} \cite{dong2022acmmm,dong2023iccv,gmmformer} was proposed as a sub-task of T2VR to fit T2VR for practical scenarios.
Previous studies tackled this task by using fine-grained matching between text queries and frame-level dense representations.
Dong \etal \ \cite{dong2022acmmm} proposed multi-scale similarity learning (MS-SL), which constructs multiple clips from encoded frame-level representations as candidate moments and computes the cross-modal similarity between the clips and queries.
In \cite{dong2023iccv}, inspired by VLM capabilities, they proposed dual learning with dynamic knowledge distillation \cite{knowledge_distill} (DL-DKD), where student branches learn the cross-modal similarity distribution between the frames and query from the teacher branch of CLIP-B/32.
These studies did not test large VLMs because of their high computation costs when extracting frame representations. This study addresses this problem by using super images to reduce the number of encodings, mitigating the low efficiency of large VLMs.
\\\textbf{Vision-language models (VLMs)} \cite{radford2021icml,fang2023cvpr,miech2020cvpr,bain2021iccv,xu2021emnlp,lei2021cvpr,ging2020neurips,yu2022coca} have gained significant attention from researchers with their powerful multimodal understanding. As image-text VLMs \cite{radford2021icml,fang2023cvpr,yu2022coca} have strong capabilities of zero-shot text-to-image retrieval performance, researchers have made tremendous efforts in training video-text VLMs \cite{miech2020cvpr,bain2021iccv,xu2021emnlp,lei2021cvpr,ging2020neurips}.
They reported that these VLMs also demonstrated exceptional performance in T2VR tasks, with larger models achieving even higher performance. However, no previous research proposed an efficient and effective method using large VLMs for PRVR. This study is the first attempt to achieve an efficient and effective PRVR with large VLMs.
\\\textbf{Frame selection/sampling} \cite{Wu_2019_CVPR,Zhi_2021_ICCV,lei2021cvpr,lei2023acl,wu2020MVFNet,wang2021efficient,ziyi2022eccv,rodriguez-opazo-etal-2023-memory,cret,Xian_2024_WACV} reduces the number of frames to mitigate the computation costs. Previous work is further categorized into two types in terms of frame encoding: (a) encode-then-reduce and (b) reduce-then-encode. While (a) focuses on selecting meaningful features \textit{after} extracting dense frame representations \cite{Wu_2019_CVPR,wu2020MVFNet,wang2021efficient,ziyi2022eccv,rodriguez-opazo-etal-2023-memory,cret}, (b) reduces the number of frames \textit{before} encoding them \cite{Zhi_2021_ICCV,lei2021cvpr,lei2023acl,Xian_2024_WACV}. When utilizing large VLMs, encoding all of the frames is costly; thus, (a) is undesirable.
Super images belong to (b), and we compare our approach with other (b) methods in \secref{subsec:comparison_to_other_approaches}.
\\\textbf{Summarized video representations} \cite{davis1997iccv,bilen2016cvpr,qiu2021iccv,tavakolian2019iccv,tavakolian2019arxiv,fan2022iclr,xu2023iccv} are another direction for efficient video processing. They focus on extracting compressed video features without compromising the essential information for downstream tasks.
The mainstream is to represent videos as single images by summarizing RGB and motion in a video \cite{bilen2016cvpr,qiu2021iccv,tavakolian2019iccv,tavakolian2019arxiv}.
Instead of representing videos as single images, super images for action recognition (SIFAR) \cite{fan2022iclr} uses super images for efficient action recognition. SIFAR represents videos as single super images and fine-tunes Swin Transformer \cite{liu2021iccv} pre-trained on ImageNet \cite{deng2009cvpr} on them.
SIFAR reported a comparable performance to SOTA methods on action recognition while reducing the computation costs significantly. Xu \etal \cite{xu2023iccv} applied the same scheme to deep fake detection and reported the effectiveness of super images.
Standing on the shoulders of these studies, this work introduces three novel contributions. First, it is the first attempt to apply super images to vision-language tasks while prior work focused solely on vision tasks. Second, we conduct zero-shot experiments and observe the strong generalization capabilities of VLMs to super images for PRVR. Last, we propose a hybrid approach using high- and low-efficiency models achieved by adjusting grid sizes and VLM backbones. This approach strikes a good balance between computation costs and retrieval performance.

\section{Approach}
In this section, we describe our zero-shot, fine-tuning, and hybrid QASIR in \secref{subsec:zero_shot_approach}, \ref{subsec:fine_tune_approach}, and \ref{subsec:reranking} after introducing the background on super images in \secref{subsec:preliminary}.
We note that the novelty of our work is to combine VLMs and super images and construct a hybrid QASIR for efficient PRVR. The network architecture itself is simple and draws from established methodologies.

\subsection{Preliminary: super images}
\label{subsec:preliminary}

\begin{figure}[t]
  \centering
  \includegraphics[width=0.8\linewidth]{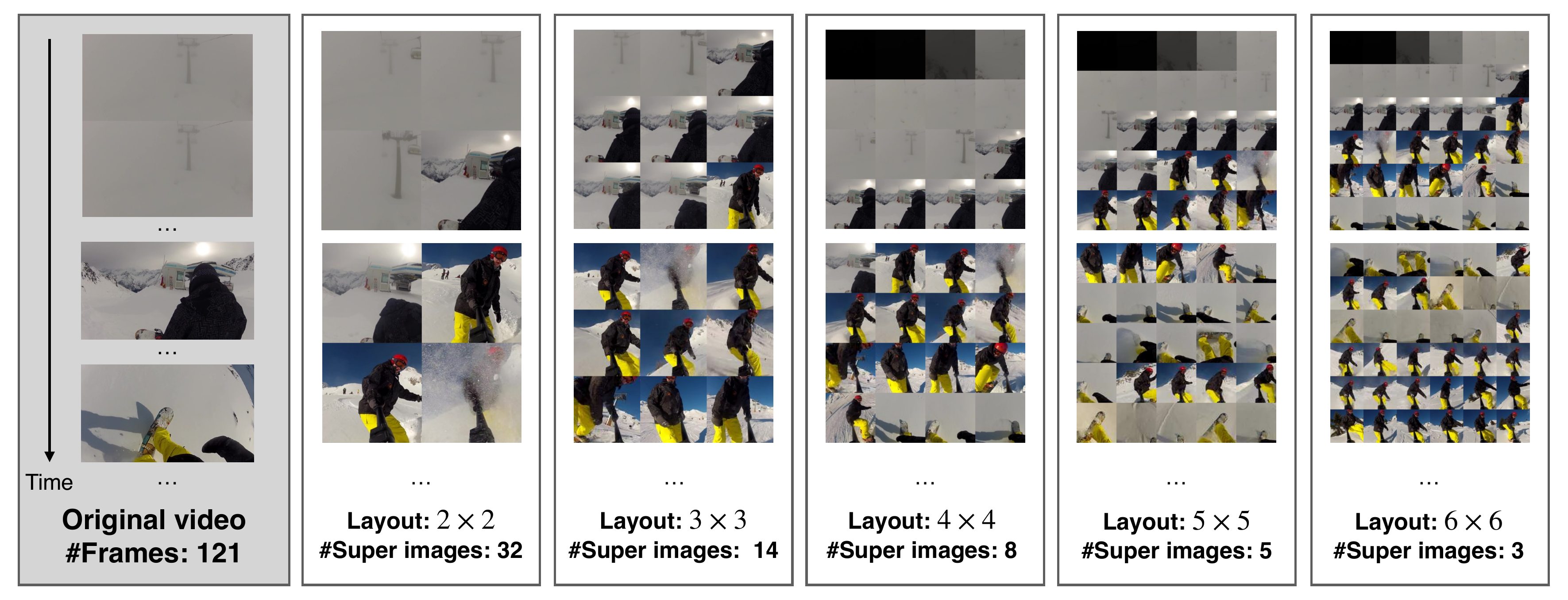}
  \caption{Comparison of original video frames and created super images in different layouts: $2\times2$, $3\times3$, $4\times4$, $5\times5$, and $6\times6$. Note that super images are placed onto grid in up-to-down and left-to-right order.}
  \label{fig:super_images}
\end{figure}

Super images were first proposed in \cite{fan2022iclr} for action recognition.
Given sequential video frames, they are ordered by a pre-defined layout to form a super image (\figref{fig:super_images}).
They reported that among various layouts, the square layout achieved the best performance across different datasets because it may encourage the models to learn temporal relationships between frames. As with SIFAR, this work uses the same layout.
To create super images, SIFAR first chooses the number of sampling frames $M$, samples frames uniformly, and forms a super image by placing them in up-to-down and left-to-right order onto a grid, either in as $(N-1) \times N$ when $M < (N-1) \times N$ or as $N \times N$ when $M \geq (N-1) \times N$, where $N = \lceil \sqrt{M} \rceil$. Empty images are padded at the end if the grid is not full.

SIFAR sparsely samples frames by choosing $M$ beforehand and creates single super images that capture the global content of videos.
However, it may be sub-optimal for PRVR because it requires models to capture local moments in the videos, and single super images via sparse sampling likely miss them.
We thus set the frame per second, sample frames, and create sequential super images by fitting them into the pre-defined grid layout.
\figref{fig:super_images} shows a comparison of the original video frames and created sequential super images. We can observe that when increasing the grid size $N$, the number of super images decreases quadratically, and objects in each grid become smaller.
In \secref{subsec:zero_shot_performance}, we delve into this matter, exploring the optimal grid size that enables models to achieve high performance efficiently.

\subsection{Zero-shot QASIR}
\label{subsec:zero_shot_approach}

\begin{figure}[t]
  \centering
  \includegraphics[width=0.88\linewidth]{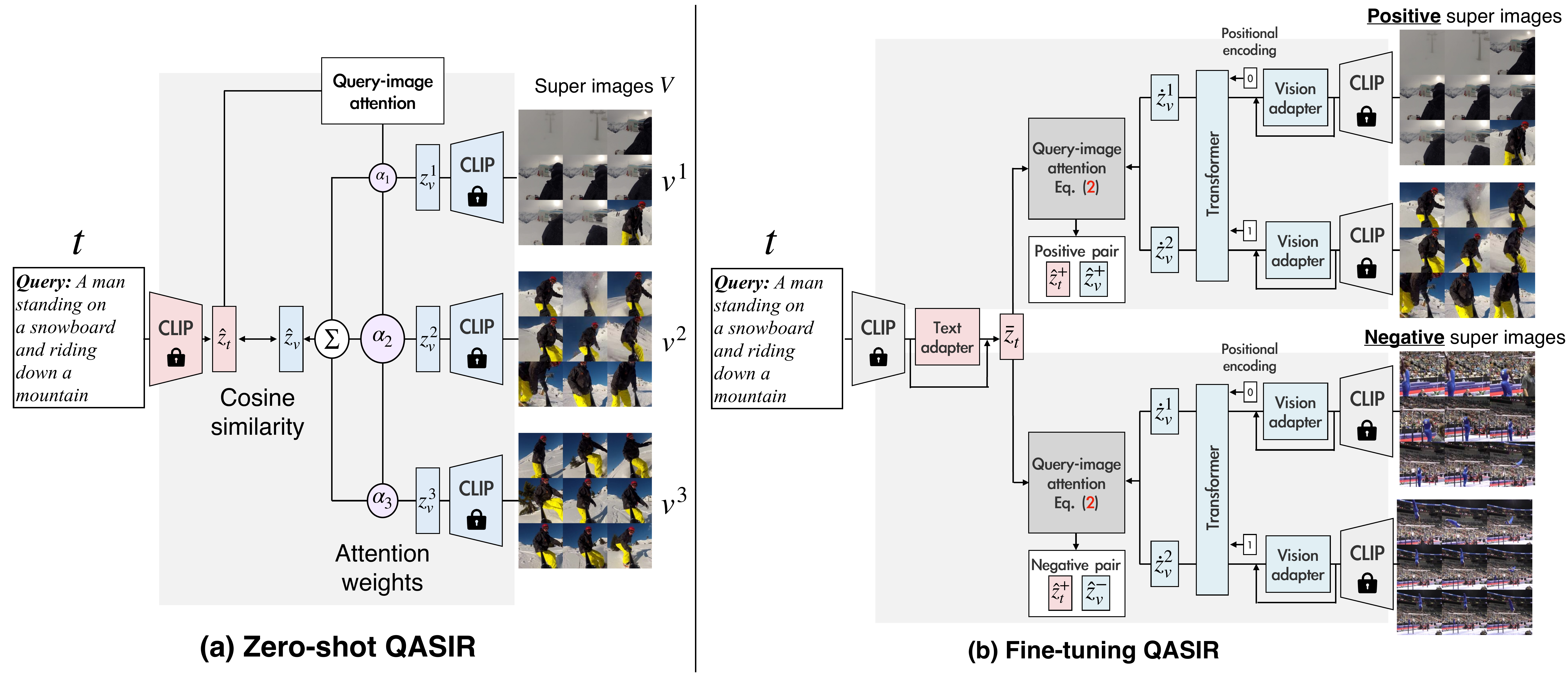}
  \caption{(a) Overview of zero-shot QASIR. (b) Fine-tuning QASIR. Given textual query $\Bdma{\hat{z}}_t$, model computes $\Bdma{\hat{z}}_t$-weighted super image vectors for positive $\Bdma{\hat{z}}_v^+$ and negative $\Bdma{\hat{z}}_v^-$ pairs. Then, their cosine similarity $\cos(\Bdma{\hat{z}}_v^+,\Bdma{\hat{z}}_t)$ and $\cos(\Bdma{\hat{z}}_v^-,\Bdma{\hat{z}}_t)$ is used for loss calculation.}
  \label{fig:qasir}
\end{figure}

In this setting, we tackle the task of retrieving super images from a corresponding partially relevant query without fine-tuning.
Let $(\Bdma{V},\Bdma{t})$ be positive pairs in the test set, where $\Bdma{V}=(\Bdma{v}^1, \Bdma{v}^2, \ldots, \Bdma{v}^k, \ldots, \Bdma{v}^K)$ and $\Bdma{t}$ represent super images and text query, respectively  ($K$ is the number of super images).
Given $\Bdma{t}$, our task is to retrieve corresponding $\Bdma{V}$ from a collection of videos, yet partial images correspond to a query in PRVR, rather than matching the whole content of $\Bdma{V}$ and $\Bdma{t}$.
Therefore, globally aggregating $\Bdma{V}$ does not perform well. We verify this through the experiments in \secref{subsec:zero_shot_performance}.

To address this problem, this study proposes a simple yet effective method, QASIR, which aggregates $\Bdma{V}$ by attending to $\Bdma{t}$ to compute similarity with an attention mechanism \cite{bahdanau2015iclr} (\figref{fig:qasir} (a)).
Specifically, the vision and text branches of VLMs convert super images $\Bdma{V}$ and a query $\Bdma{t}$ into their latent representations $\Bdma{Z}_v \in \mathbb{R}^{K \times d}$ and $\Bdma{z}_t \in \mathbb{R}^d$ as:
\begin{eqnarray}
    \Bdma{Z}_v = (\Bdma{z}_v^1, \Bdma{z}_v^2, \ldots, \Bdma{z}_v^k, \ldots, \Bdma{z}_v^K),\  \Bdma{z}_v^k = \mathrm{VLM_v}(\Bdma{v}^k),\ \Bdma{z}_t = \mathrm{VLM_t}(\Bdma{t}),
\end{eqnarray}
where $\mathrm{VLM}_v(\cdot),\mathrm{VLM}_t(\cdot)$ are the vision and text branches of VLMs, and $d$ represents the dimension of the latent vectors.
Then, the super image representations are aggregated into the query-attentive visual vectors $\hat{\Bdma{z}}_v$ as follows:
\begin{eqnarray}
\label{eq:query_attention_1}
    \alpha_k = \frac{\exp{( \Bdma{z}_v^k \cdot \Bdma{z}_t )}}{\sum_{i}\exp{( \Bdma{z}_v^i\cdot\Bdma{z}_t )}},\ \hat{\Bdma{z}}_v = \sum_{k}\alpha_k \Bdma{z}_v^k.
\end{eqnarray}
The similarity scores are computed by cosine similarity between $\hat{\Bdma{z}}_v$ and $\Bdma{z}_t$.

\subsection{Fine-tuning QASIR}
\label{subsec:fine_tune_approach}

Based on the zero-shot QASIR, we propose a fine-tuning QASIR to boost the retrieval performance without harming the computational efficiency (\figref{fig:qasir} (b)).
We add two modules to QASIR: feature adapters and temporal encoder.

The feature adapters are based on the CLIP adapter \cite{gao2023ijcv}, which freezes the original VLM networks and trains additional small networks, e.g., multi-layered (typically two or three) perceptrons (MLPs) with activation functions.
The researchers reported that compared with training all of the layers in VLMs, adaptation-based methods can achieve promising performance with minimal training costs, preventing the VLMs from catastrophic forgetting \cite{houlsby2019icml}.
We first describe the vision branch.
The vision adapter computes $\bar{\Bdma{z}}_v^k$ by interpolating $\Bdma{z}_v^k$ and the output vectors from MLPs as $\bar{\Bdma{z}}_v^k = \beta_v \mathrm{MLP}_v(\Bdma{z}_v^k) + (1 - \beta_v)\Bdma{z}_v^k$, where $\mathrm{MLP}_v(\cdot)$ represents the three-layer MLPs with ReLU functions, and $\beta_v$ is a hyper parameter.
Similarly, we prepare a text adapter, which computes $\bar{\Bdma{z}}_t = \beta_t \mathrm{MLP}_t(\Bdma{z}_t) + (1-\beta_t) \Bdma{z}_t$, where $\mathrm{MLP}_t(\cdot)$ is another MLP that has the same structure as $\mathrm{MLP}_v(\cdot)$.
The MLP architecture is written in Appendix A.

The temporal encoder represents the temporal information of super images. To this end, we add a single-layer Transformer \cite{vaswani2017neurips} to the vision side.
Based on the super image vectors $\bar{\Bdma{z}}_v^k$ from the vision adapter, we add positional encoding (PE) and input them into the Transformer as $\dot{\Bdma{z}}_v^k = \mathrm{Transformer}(\bar{\Bdma{z}}_v^k + \mathrm{PE}(k))$,
where $\mathrm{Transformer(\cdot)}$ and $\mathrm{PE}(\cdot)$ represent the Transformer layer and sinusoidal PE.
Then, we acquire the query-attentive visual vectors $\hat{\Bdma{z}}_v^k$ by replacing $\Bdma{z}_v^k$ with $\dot{\Bdma{z}}_v^k$ in \eqref{eq:query_attention_1}. The similarity scores are computed between $\hat{\Bdma{z}}_v^k$ and $\bar{\Bdma{z}}_t$.
\\\textbf{Loss calculation.}
To fine-tune the model, as with CLIP, we use a symmetric cross-entropy loss over the similarity scores.
From an anchor query vector $\hat{\Bdma{z}}_t^+$, let the positive and negative super image vectors be $\hat{\Bdma{z}}_v^+, \hat{\Bdma{z}}_v^-$.
Conversely, from an anchor super image vector $\hat{\Bdma{z}}_v^+$, let the positive and negative query vectors be $\hat{\Bdma{z}}_t^+, \hat{\Bdma{z}}_t^-$.
The total loss of $\mathcal{L}=\mathcal{L}_q + \mathcal{L}_i$ is computed as follows:
\begin{eqnarray}
{\footnotesize \mathcal{L}_{q}}&=&{\footnotesize \frac{1}{|\mathcal{B}|}\!\!\!\!\!\sum_{(\Bdma{z}_v^+,\Bdma{z}_t^+) \in \mathcal{B}}\!\!\!\!\!\!\!\!\log{\mleft\{\frac{\cos{(\Bdma{z}_v^+,\Bdma{z}_t^+)}}{\cos{(\Bdma{z}_v^+,\Bdma{z}_t^+)}+\sum_{\Bdma{z}_v^- \in \mathcal{N}}\cos{(\Bdma{z}_v^-,\Bdma{z}_t^+)}}\mright\}}}, \\
{\footnotesize \mathcal{L}_{i}}&=&{\footnotesize \frac{1}{|\mathcal{B}|}\!\!\!\!\!\sum_{(\Bdma{z}_v^+,\Bdma{z}_t^+) \in \mathcal{B}}\!\!\!\!\!\!\!\!\log{\mleft\{\frac{\cos{(\Bdma{z}_v^+,\Bdma{z}_t^+)}}{\cos{(\Bdma{z}_v^+,\Bdma{z}_t^+)}+\sum_{\Bdma{z}_t^- \in \mathcal{N}}\cos{(\Bdma{z}_v^+,\Bdma{z}_t^-)}}\mright\}}},
\end{eqnarray}
where $\mathcal{L}_{q}, \mathcal{L}_{i}$ represents query-to-image/image-to-query losses, and $\mathcal{B},\mathcal{N}$ indicates the samples from the mini-batch and negatives, respectively.

\subsection{Hybrid QASIR of high- and low-efficiency models}
\label{subsec:reranking}
The trade-off parameters of QASIR are the grid size, VLM size, and image resolutions between the efficiency and retrieval performance.
We confirm this through the zero-shot experiments in \secref{subsec:zero_shot_performance}.
To strike a balance between them, we propose constructing a hybrid QASIR by using high- (larger $N$ and small VLMs) and low- (smaller $N$ and large VLMs) efficiency models.
Specifically, as with \cite{miech2021cvpr}, we initially use a high-efficiency QASIR to quickly screen out irrelevant videos based on the input query then re-rank the top $R$ videos using the low-efficiency QASIR to refine the retrieval results. Note that $R$ is a hyper-parameter, which is substantially smaller than the number of total videos in the dataset.

\section{Experiments}
\label{sec:experiments}

\begin{table*}[t]
\centering
\caption{Comparison of models. DL-DKD uses CLIP-B/32$^\ast$ only in training phase to transfer VLM knowledge into visual and textual backbone.}
\scalebox{0.5}{
\begin{tabular}{c|c|c|cc|c|c}
\hline
& Model name & Resolution & Vision backbone & Text backbone & Parameters (M) & Image-level GFLOPs \\ \hline
& MS-SL & 224 & I3D (+ResNet152) & RoBERTa & 433.3 & 40.0 \\
& GMMFormer & 224 & I3D (+ResNet152) & RoBERTa & 441.3 & 40.0 \\
& DL-DKD & 224 & I3D (+ResNet152, CLIP-B/32$^\ast$) & RoBERTa$^\dagger$ (+CLIP-B/32$^\ast$) & 439.0 & 40.0 (+8.8$^\ast$) \\
\multirow{-4}{*}{lightweight SOTA} & MS-SL (B32) & 224 & \multicolumn{2}{c|}{CLIP-B/32} & 156.3 & 8.8 \\ \hline
& MS-SL (L14) & 224 & \multicolumn{2}{c|}{CLIP-L/14} & 432.6 & 162.0 \\
\multirow{-2}{*}{heavyweight SOTA} & MS-SL (L14-336) & 336 & \multicolumn{2}{c|}{CLIP-L/14-336} & 432.9 & 381.9 \\ \hline
& QASIR-B32 & 224 & \multicolumn{2}{c|}{CLIP-B/32} & 151.3 & 8.8 \\
& QASIR-L14 & 224 & \multicolumn{2}{c|}{CLIP-L/14} & 427.6 & 162.0 \\
\multirow{-3}{*}{Zero-shot} & QASIR-L14-336 & 336 & \multicolumn{2}{c|}{CLIP-L/14-336} & 427.9 & 381.9 \\ \hline
& QASIR-B32 & 224 & \multicolumn{2}{c|}{CLIP-B/32} & 154.7 & 8.8 \\
& QASIR-L14 & 224 & \multicolumn{2}{c|}{CLIP-L/14} & 431.0 & 162.0 \\
\multirow{-3}{*}{Fine-tuned} & QASIR-L14-336 & 336 & \multicolumn{2}{c|}{CLIP-L/14-336} &  431.3 & 381.9 \\ \hline
\end{tabular}
}
\label{tab:model_comparison}
\end{table*}

\subsection{Experimental settings}

\textbf{Datasets.} We evaluate our methods on three PRVR benchmark datasets: ActivityNet Captions \cite{krishna2017iccv}, TVR \cite{lei2020eccv}, and Charades-STA \cite{gao2017iccv}. We follow the same split as the previous studies \cite{dong2022acmmm,dong2023iccv}. Detailed dataset statistics are provided in Appendix B.
\\\textbf{Evaluation metrics.}
We prepare evaluation metrics in terms of both performance and computation cost. As performance metrics, following previous PRVR studies \cite{dong2022acmmm,dong2023iccv}, we use Recall@$K$ ($K=1,5,10,100$), which computes the percentage of queries that correctly retrieve positive videos in the top $K$ of the ranking list. We also report the sum of all recalls (sumR) for overall comparison. As computation cost metrics, we report the number of parameters (\#parameters) and video-text GFLOPs, which computes the total number of floating point operations from visual/textual backbone encodings to video-text similarity calculation. The definition of video-text GFLOPs is provided in Appendix C.
\\\textbf{SOTA models.}
\tabref{tab:model_comparison} shows a comparison of models. We compare our approaches with three SOTA methods that reported the best scores on the PRVR task: MS-SL \cite{dong2022acmmm}, GMMFormer \cite{gmmformer}, and DL-DKD \cite{dong2023iccv}.
They use the same backbones, I3D \cite{carreira2017cvpr}/ResNet152 \cite{he2016cvpr} and RoBERTa \cite{liu2019roberta}, for the visual and textual sides.
We provide detailed settings on the conventional backbones in Appendix D.
In addition, to investigate the impact of VLM backbones, we prepare additional MS-SL models whose backbone was substituted with CLIP-B/32, CLIP-L/14, and CLIP-L/14-336.
We do not use GMMFormer/DL-DKD instead of MS-SL because (1) MS-SL with VLMs slightly achieves better than GMMFormer with VLMs in our preliminary experiments and (2) DL-DKD requires teacher and student branches but it is unexpected that the student (L32, L14, L14/336) is equivalent or superior to the teacher (B/32).
Note that CLIP-L/14 and CLIP-L/14-336 have substantially higher GFLOPs than CLIP-B/32 when encoding an image.
To differentiate between SOTA models utilizing lightweight backbones and heavyweight ones, we term the former \textit{lightweight SOTA} and the latter \textit{heavyweight SOTA}.
\\\textbf{Grid sizes, image resolution, and VLM backbones.}
To investigate how the grid sizes, image resolution, and VLM backbones affect the retrieval performance and computation costs of PRVR, we prepare the three zero-shot and fine-tuned models listed in \tabref{tab:model_comparison}.
A comparison of different VLMs tests whether the model size of the VLMs affects the PRVR performance.
In addition, a comparison with an input resolution of 224/336 and grid sizes of $1\times1$ to $6\times6$ tests whether the visibility of objects in super images affects the performance. Note that $1\times1$ is equivalent to frame-level representations. Other implementation details are described in Appendix E.

\subsection{Zero-shot evaluation}
\label{subsec:zero_shot_performance}

\begin{table*}[t]
\centering
\caption{Zero-shot QASIR performance on benchmark datasets. \textbf{Bold} values indicate highest scores among each QASIR model.}
\scalebox{0.48}{
\begin{tabular}{c|ccccccccccccccccccccc}
\hline
\rowcolor[HTML]{EFEFEF} 
 & \multicolumn{7}{c|}{\cellcolor[HTML]{EFEFEF}ActivityNet Captions} & \multicolumn{7}{c|}{\cellcolor[HTML]{EFEFEF}TVR} & \multicolumn{7}{c}{\cellcolor[HTML]{EFEFEF}Charades-STA} \\ \hline
& \#Frames & GFLOPs & R@1 & R@5 & R@10 & R@100 & \multicolumn{1}{c|}{sumR} & \#Frames & GFLOPs & R@1 & R@5 & R@10 & R@100 & \multicolumn{1}{c|}{sumR} & \#Frames & GFLOPs & R@1 & R@5 & R@10 & R@100 & \multicolumn{1}{c}{sumR} \\ \hline
MS-SL & 118.2 & 3.4$\times$10$^3$ & 7.1 & 22.5 & 34.7 & 75.8 & \multicolumn{1}{c|}{140.1} & 1143.0 & 4.6$\times$10$^4$ & 13.5 & 32.1 & 43.4 & 83.4 & \multicolumn{1}{c|}{172.4} & 31.1 & 1.3$\times$10$^3$ & 1.8 & 7.1 & 11.8 & 47.7 & 68.4 \\
GMMFormer & 118.2 & 3.4$\times$10$^3$ & 8.3 & 24.9 & 36.7 & 76.1 & \multicolumn{1}{c|}{146.0} & 1143.0 & 4.6$\times$10$^4$ & 13.9 & 33.3 & 44.5 & 84.9 & \multicolumn{1}{c|}{176.6} & 31.1 & 1.3$\times$10$^3$ & 2.1 & 7.8 & 12.5 & 50.6 & 72.9 \\
DL-DKD & 118.2 & 3.4$\times$10$^3$ & 8.0 & 25.0 & 37.5 & 77.1 & \multicolumn{1}{c|}{147.6} & 1143.0 & 4.6$\times$10$^4$ & 14.4 & 34.9 & 45.8 & 84.9 & \multicolumn{1}{c|}{179.9} & - & - & - & - & - & - & - \\
\rowcolor[HTML]{EFEFEF}
\multicolumn{22}{l}{\cellcolor[HTML]{EFEFEF}QASIR-B32} \\
1$\times$1 & 60.3 & 5.4$\times$10$^2$ & \textbf{11.1} & \textbf{26.4} & \textbf{35.9} & \textbf{70.9} & \multicolumn{1}{c|}{\textbf{144.4}} & 229.4 & 2.0$\times$10$^3$ & \textbf{9.5} & \textbf{21.2} & \textbf{28.2} & \textbf{61.1} & \multicolumn{1}{c|}{\textbf{120.0}} & 31.1 & 2.8$\times$10$^2$ & \textbf{1.3} & \textbf{5.1} & \textbf{8.4} & \textbf{36.4} & \textbf{51.1} \\
2$\times$2 & 15.5 & 1.4$\times$10$^2$ & 10.3 & 25.0 & 34.5 & 69.1 & \multicolumn{1}{c|}{138.9} & 57.7 & 5.1$\times$10$^2$ & 7.3 & 17.2 & 23.7 & 57.3 & \multicolumn{1}{c|}{105.5} & 8.1 & 7.7$\times$10$^1$ & 1.0 & 4.3 & 6.7 & 30.3 & 42.3 \\
3$\times$3 & 7.1 & 6.9$\times$10$^1$ & 8.3 & 21.1 & 29.4 & 64.3 & \multicolumn{1}{c|}{123.1} & 23.9 & 2.3$\times$10$^2$ & 4.9 & 12.1 & 17.4 & 50.8 & \multicolumn{1}{c|}{85.1} & 3.9 & 4.0$\times$10$^1$ & 1.0 & 3.0 & 4.8 & 22.2 & 30.9 \\
4$\times$4 & 4.2 & 4.3$\times$10$^1$ & 6.2 & 16.9 & 24.2 & 56.7 & \multicolumn{1}{c|}{104.0} & 14.8 & 1.3$\times$10$^2$ & 3.1 & 8.7 & 12.8 & 42.9 & \multicolumn{1}{c|}{67.5} & 2.4 & 2.7$\times$10$^1$ & 0.5 & 1.6 & 3.0 & 17.8 & 23.0 \\
5$\times$5 & 2.9 & 3.1$\times$10$^1$ & 4.1 & 12.2 & 18.3 & 47.2 & \multicolumn{1}{c|}{81.9} & 9.6 & 9.0$\times$10$^1$ & 1.8 & 5.8 & 9.2 & 34.4 & \multicolumn{1}{c|}{51.2} & 1.8 & 2.1$\times$10$^1$ & 0.2 & 1.1 & 1.8 & 14.4 & 17.5 \\
6$\times$6 & 2.1 & 2.5$\times$10$^1$ & 3.1 & 9.3 & 13.7 & 38.8 & \multicolumn{1}{c|}{64.8} & 6.8 & 6.6$\times$10$^1$ & 1.0 & 3.4 & 5.6 & 26.8 & \multicolumn{1}{c|}{36.8} & 1.1 & 1.5$\times$10$^1$ & 0.2 & 0.9 & 1.7 & 12.8 & 15.6 \\
\rowcolor[HTML]{EFEFEF}
\multicolumn{22}{l}{\cellcolor[HTML]{EFEFEF}QASIR-L14} \\
1$\times$1 & 60.3 & 9.8$\times$10$^3$ & \textbf{14.2} & \textbf{31.7} & \textbf{42.7} & 77.0 & \multicolumn{1}{c|}{\textbf{165.6}} & 229.4 & 3.7$\times$10$^4$ & \textbf{16.4} & \textbf{32.5} & \textbf{40.7} & \textbf{74.0} & \multicolumn{1}{c|}{\textbf{163.6}} & 31.1 & 5.0$\times$10$^3$ & \textbf{2.7} & \textbf{8.3} & \textbf{13.3} & \textbf{48.6} & \textbf{72.8} \\
2$\times$2 & 15.5 & 2.5$\times$10$^3$ & 13.7 & 31.3 & 41.9 & \textbf{77.1} & \multicolumn{1}{c|}{164.0} & 57.7 & 9.2$\times$10$^3$ & 14.1 & 29.6 & 37.4 & 72.5 & \multicolumn{1}{c|}{154.0} & 8.1 & 1.3$\times$10$^3$ & 2.0 & 6.9 & 11.2 & 43.7 & 63.8 \\
3$\times$3 & 7.1 & 1.1$\times$10$^3$ & 12.4 & 29.4 & 39.5 & 75.0 & \multicolumn{1}{c|}{156.4} & 23.9 & 4.1$\times$10$^3$ & 11.8 & 25.4 & 33.2 & 68.9 & \multicolumn{1}{c|}{139.3} & 3.9 & 6.4$\times$10$^2$ & 1.7 & 5.4 & 9.0 & 36.2 & 52.3 \\
4$\times$4 & 4.2 & 7.0$\times$10$^2$ & 10.7 & 26.0 & 36.1 & 70.8 & \multicolumn{1}{c|}{143.8} & 14.8 & 2.3$\times$10$^3$ & 8.4 & 19.0 & 26.3 & 61.6 & \multicolumn{1}{c|}{115.4} & 2.4 & 4.0$\times$10$^2$ & 1.0 & 4.6 & 7.4 & 31.1 & 44.2 \\
5$\times$5 & 2.9 & 4.8$\times$10$^2$ & 9.0 & 23.2 & 32.5 & 67.3 & \multicolumn{1}{c|}{132.0} & 9.6 & 1.5$\times$10$^3$ & 6.0 & 14.3 & 20.0 & 54.8 & \multicolumn{1}{c|}{95.1} & 1.8 & 3.0$\times$10$^2$ & 1.0 & 3.4 & 5.6 & 25.8 & 35.8 \\
6$\times$6 & 2.1 & 3.7$\times$10$^2$ & 7.7 & 19.8 & 28.2 & 62.4 & \multicolumn{1}{c|}{118.0} & 6.8 & 1.1$\times$10$^3$ & 4.6 & 12.1 & 16.9 & 49.2 & \multicolumn{1}{c|}{82.9} & 1.1 & 1.9$\times$10$^2$ & 0.9 & 3.0 & 4.9 & 23.3 & 32.2 \\
\rowcolor[HTML]{EFEFEF}
\multicolumn{22}{l}{\cellcolor[HTML]{EFEFEF}QASIR-L14-336} \\
1$\times$1 & 60.3 & 2.3$\times$10$^4$ & 14.6 & 32.8 & 43.8 & 77.9 & \multicolumn{1}{c|}{169.1} & 229.4 & 8.7$\times$10$^4$ & 16.8 & \textbf{34.1} & \textbf{42.5} & 75.4 & \multicolumn{1}{c|}{\textbf{168.8}} & 31.1 & 1.1$\times$10$^4$ & \textbf{2.6} & \textbf{8.5} & \textbf{13.8} & \textbf{51.0} & \textbf{75.8} \\
2$\times$2 & 15.5 & 5.9$\times$10$^3$ & \textbf{15.7} & \textbf{33.9} & \textbf{45.3} & \textbf{78.7} & \multicolumn{1}{c|}{\textbf{173.6}} & 57.7 & 2.1$\times$10$^4$ & \textbf{17.2} & 33.3 & 42.1 & \textbf{76.1} & \multicolumn{1}{c|}{168.7} & 8.1 & 3.1$\times$10$^3$ & 2.3 & 7.8 & 13.2 & 48.1 & 71.5 \\
3$\times$3 & 7.1 & 2.7$\times$10$^3$ & 14.8 & 32.3 & 44.0 & 77.5 & \multicolumn{1}{c|}{169.6} & 23.9 & 9.8$\times$10$^3$ & 15.2 & 30.9 & 39.3 & 73.7 & \multicolumn{1}{c|}{159.1} & 3.9 & 1.4$\times$10$^3$ & 1.6 & 6.9 & 11.5 & 44.1 & 64.1 \\
4$\times$4 & 4.2 & 1.6$\times$10$^3$ & 13.2 & 30.8 & 41.3 & 75.4 & \multicolumn{1}{c|}{160.6} & 14.8 & 5.6$\times$10$^3$ & 12.4 & 26.6 & 34.5 & 69.9 & \multicolumn{1}{c|}{143.4} & 2.4 & 9.3$\times$10$^2$ & 1.7 & 5.8 & 9.4 & 37.2 & 54.1 \\
5$\times$5 & 2.9 & 1.1$\times$10$^3$ & 11.9 & 28.1 & 38.5 & 73.0 & \multicolumn{1}{c|}{151.8} & 9.6 & 3.6$\times$10$^3$ & 9.9 & 22.0 & 29.2 & 65.5 & \multicolumn{1}{c|}{126.6} & 1.8 & 7.0$\times$10$^2$ & 1.3 & 4.8 & 8.0 & 32.9 & 47.0 \\
6$\times$6 & 2.1 & 8.4$\times$10$^2$ & 10.6 & 25.7 & 35.5 & 70.1 & \multicolumn{1}{c|}{141.9} & 6.8 & 2.6$\times$10$^3$ & 7.7 & 18.0 & 24.5 & 59.8 & \multicolumn{1}{c|}{110.0} & 1.1 & 4.4$\times$10$^2$ & 1.3 & 4.0 & 6.6 & 28.0 & 39.9 \\
\hline
\end{tabular}
}
\label{tab:zero_shot_all}
\end{table*}

\noindent
\textbf{VLMs generalize well to super images.} \tabref{tab:zero_shot_all} shows the zero-shot performance on the benchmark datasets. Surprisingly, zero-shot QASIR with super images achieves promising performance against the lightweight SOTA methods.
In addition, the $2\times2$ models minimize the retrieval performance decrease while reducing the computation costs to $\frac{1}{4}$ from $1\times1$.
\\\textbf{Grid size, image resolution, and VLM size are trade-off parameters between performance and computation costs.} As mentioned in \secref{subsec:preliminary}, the grid size and image resolution are related to the visibility of objects. A higher image resolution and smaller grid sizes contribute to enhanced retrieval performance. However, such improvements come at the expense of increased computation costs. Moreover, larger VLMs demonstrate superior performance, albeit at increased computation costs.
\\\textbf{Performance difference across datasets.} QASIR consistently achieves higher performance than the SOTA methods that use conventional ResNet + RoBERTa backbones on ActivityNet Captions and Charades-STA but falls short on TVR. We notice that most nouns in queries are general objects (e.g., ``man'' and ``dog'') in ActivityNet Captions and Charades-STA, while they are character names (e.g., ``Sheldon'') in TVR. Therefore, one possible reason for the gap is that the VLMs do not know these TV show characters and fail to link the character names and their visual representations. See Appendix F for a detailed discussion.
\\\textbf{Is query attention best?}
To verify the effectiveness of query attention, we compare it with other aggregation methods: mean and max pooling. We evaluate the zero-shot performance using $2\times2$ QASIR-L-14-336 on ActivityNet Captions. As a result, query attention (sumR$=$173.6) substantially outperforms mean (144.1) and max (120.2). This indicates that attending to local segments in super images for an input query is crucial for PRVR.

\subsection{Fine-tuning evaluation}
\begin{table*}[t]
\centering
\caption{Fine-tuning QASIR performance on benchmark datasets.}
\scalebox{0.57}{
\begin{tabular}{lcccccccccccccccccc}
\hline
\rowcolor[HTML]{EFEFEF} 
\multicolumn{1}{l|}{\cellcolor[HTML]{EFEFEF}} & \multicolumn{6}{c|}{\cellcolor[HTML]{EFEFEF}ActivityNet Captions} & \multicolumn{6}{c|}{\cellcolor[HTML]{EFEFEF}TVR} & \multicolumn{6}{c}{\cellcolor[HTML]{EFEFEF}Charades-STA} \\ \hline
\multicolumn{1}{c|}{} & GFLOPs & R@1 & R@5 & R@10 & R@100 & \multicolumn{1}{c|}{sumR} & GFLOPs & R@1 & R@5 & R@10 & R@100 & \multicolumn{1}{c|}{sumR} & GFLOPs & R@1 & R@5 & R@10 & R@100 & sumR \\ \hline
\multicolumn{1}{c|}{MS-SL} & 3.4$\times$10$^3$ & 7.1 & 22.5 & 34.7 & 75.8 & \multicolumn{1}{c|}{140.1} & 4.6$\times$10$^4$ & 13.5 & 32.1 & 43.4 & 83.4 & \multicolumn{1}{c|}{172.4} & 1.3$\times$10$^3$ & 1.8 & 7.1 & 11.8 & 47.7 & 68.4 \\
\multicolumn{1}{c|}{GMMFormer} & 3.4$\times$10$^3$ & 8.3 & 24.9 & 36.7 & 76.1 & \multicolumn{1}{c|}{146.0} & 4.6$\times$10$^4$ & 13.9 & 33.3 & 44.5 & 84.9 & \multicolumn{1}{c|}{176.6} & 1.3$\times$10$^3$ & 2.1 & 7.8 & 12.5 & 50.6 & 72.9 \\
\multicolumn{1}{c|}{DL-DKD} & 3.4$\times$10$^3$ & 8.0 & 25.0 & 37.5 & 77.1 & \multicolumn{1}{c|}{147.6} & 4.6$\times$10$^4$ & 14.4 & 34.9 & 45.8 & 84.9 & \multicolumn{1}{c|}{179.9} & - & - & - & - & - & - \\
\multicolumn{1}{c|}{MS-SL (B32)} & 5.4$\times$10$^2$ & 11.7 & 31.2 & 43.5 & 81.7 & \multicolumn{1}{c|}{168.1} & 2.0$\times$10$^3$ & 17.5 & 39.5 & 51.3 & 88.3 & \multicolumn{1}{c|}{196.6} & 2.8$\times$10$^2$ & 1.2 & 4.4 & 7.6 & 41.5 & 54.7 \\
\multicolumn{1}{c|}{MS-SL (L14)} & 9.8$\times$10$^3$ & 14.8 & 36.6 & 50.3 & 84.8 & \multicolumn{1}{c|}{186.5} & 3.7$\times$10$^4$ & 22.0 & 46.2 & 58.0 & 91.1 & \multicolumn{1}{c|}{217.3} & 5.0$\times$10$^3$ & 1.7 & 7.1 & 12.0 & 54.4 & 75.2 \\
\multicolumn{1}{c|}{MS-SL (L14-336)} & 2.3$\times$10$^4$ & 14.2 & 36.9 & 50.4 & 85.2 & \multicolumn{1}{c|}{186.7} & 8.7$\times$10$^4$ & 24.7 & 49.3 & 60.6 & 92.0 & \multicolumn{1}{c|}{226.6} & 1.1$\times$10$^4$ & 2.3 & 8.6 & 14.1 & 57.6 & 82.6 \\ \hline
\rowcolor[HTML]{EFEFEF}
\multicolumn{19}{l}{\cellcolor[HTML]{EFEFEF}QASIR-B32} \\
\multicolumn{1}{c|}{$1\times1$} & 5.4$\times$10$^2$ & \textbf{14.1} & \textbf{32.9} & \textbf{44.5} & 79.9 & \multicolumn{1}{c|}{\textbf{171.4}} & 2.0$\times$10$^3$ & \textbf{19.0} & \textbf{39.9} & \textbf{50.4} & \textbf{87.2} & \multicolumn{1}{c|}{\textbf{196.5}} & 2.8$\times$10$^2$ & \textbf{1.9} & \textbf{5.8} & \textbf{10.1} & \textbf{40.0} & \textbf{57.8} \\
\multicolumn{1}{c|}{$2\times2$} & 1.4$\times$10$^2$ & 13.2 & 31.7 & 43.1 & \textbf{80.0} & \multicolumn{1}{c|}{167.9} & 5.1$\times$10$^2$ & 15.8 & 34.7 & 45.5 & 84.5 & \multicolumn{1}{c|}{180.4} & 7.7$\times$10$^1$ & 1.3 & 4.9 & 8.6 & 35.5 & 50.3 \\
\multicolumn{1}{c|}{$3\times3$} & 6.9$\times$10$^1$ & 10.7 & 26.8 & 36.9 & 75.0 & \multicolumn{1}{c|}{149.3} & 2.3$\times$10$^2$ & 11.6 & 28.5 & 38.5 & 80.0 & \multicolumn{1}{c|}{158.7} & 4.0$\times$10$^1$ & 0.9 & 4.1 & 6.7 & 28.3 & 40.0 \\
\multicolumn{1}{c|}{$4\times4$} & 4.3$\times$10$^1$ & 8.4 & 22.5 & 32.4 & 69.7 & \multicolumn{1}{c|}{133.0} & 1.3$\times$10$^2$ & 8.6 & 22.2 & 31.5 & 74.5 & \multicolumn{1}{c|}{136.8} & 2.7$\times$10$^1$ & 0.7 & 2.9 & 4.8 & 24.6 & 32.9 \\
\multicolumn{1}{c|}{$5\times5$} & 3.1$\times$10$^1$ & 6.2 & 17.7 & 25.8 & 62.1 & \multicolumn{1}{c|}{111.9} & 9.0$\times$10$^1$ & 6.3 & 17.7 & 25.6 & 68.8 & \multicolumn{1}{c|}{118.4} & 2.1$\times$10$^1$ & 0.4 & 2.0 & 3.4 & 19.8 & 25.6 \\
\multicolumn{1}{c|}{$6\times6$} & 2.5$\times$10$^1$ & 4.8 & 14.0 & 20.8 & 55.2 & \multicolumn{1}{c|}{94.7} & 6.6$\times$10$^1$ & 4.8 & 13.7 & 21.0 & 63.3 & \multicolumn{1}{c|}{102.8} & 1.5$\times$10$^1$ & 0.5 & 1.8 & 3.1 & 18.3 & 23.7 \\
\rowcolor[HTML]{EFEFEF}
\multicolumn{19}{l}{\cellcolor[HTML]{EFEFEF}QASIR-L14} \\
\multicolumn{1}{c|}{$1\times1$} & 9.8$\times$10$^3$ & \textbf{18.9} & \textbf{41.0} & \textbf{53.2} & \textbf{84.8} & \multicolumn{1}{c|}{\textbf{197.9}} & 3.7$\times$10$^4$ & \textbf{23.6} & \textbf{47.0} & \textbf{56.5} & \textbf{89.0} & \multicolumn{1}{c|}{\textbf{216.1}} & 5.0$\times$10$^3$ & \textbf{3.0} & \textbf{11.2} & \textbf{17.4} & \textbf{55.7} & \textbf{87.2} \\
\multicolumn{1}{c|}{$2\times2$} & 2.5$\times$10$^3$ & 18.2 & 39.6 & 52.2 & \textbf{84.8} & \multicolumn{1}{c|}{194.9} & 9.2$\times$10$^3$ & 23.0 & 45.4 & 56.3 & 88.9 & \multicolumn{1}{c|}{213.6} & 1.3$\times$10$^3$ & 2.7 & 9.3 & 14.9 & 49.1 & 76.0 \\
\multicolumn{1}{c|}{$3\times3$} & 1.1$\times$10$^3$ & 16.5 & 37.5 & 49.9 & 83.2 & \multicolumn{1}{c|}{187.1} & 4.1$\times$10$^3$ & 21.0 & 42.5 & 54.0 & 88.4 & \multicolumn{1}{c|}{205.8} & 6.4$\times$10$^2$ & 2.0 & 7.9 & 12.2 & 43.3 & 65.3 \\
\multicolumn{1}{c|}{$4\times4$} & 7.0$\times$10$^2$ & 14.1 & 33.8 & 45.6 & 80.4 & \multicolumn{1}{c|}{174.0} & 2.3$\times$10$^3$ & 16.3 & 35.7 & 46.4 & 84.6 & \multicolumn{1}{c|}{182.9} & 4.0$\times$10$^2$ & 1.8 & 6.3 & 9.6 & 36.8 & 54.4 \\
\multicolumn{1}{c|}{$5\times5$} & 4.8$\times$10$^2$ & 12.5 & 30.3 & 41.8 & 78.1 & \multicolumn{1}{c|}{162.8} & 1.5$\times$10$^3$ & 13.5 & 31.7 & 42.0 & 81.6 & \multicolumn{1}{c|}{168.9} & 3.0$\times$10$^2$ & 1.3 & 4.2 & 6.9 & 31.1 & 43.4 \\
\multicolumn{1}{c|}{$6\times6$} & 3.7$\times$10$^2$ & 10.7 & 27.2 & 37.8 & 75.2 & \multicolumn{1}{c|}{150.9} & 1.1$\times$10$^3$ & 12.0 & 27.8 & 37.7 & 78.9 & \multicolumn{1}{c|}{156.4} & 1.9$\times$10$^2$ & 1.2 & 3.7 & 6.2 & 27.2 & 38.3 \\
\rowcolor[HTML]{EFEFEF}
\multicolumn{19}{l}{\cellcolor[HTML]{EFEFEF}QASIR-L14-336} \\
\multicolumn{1}{c|}{$1\times1$} & 2.3$\times$10$^4$ & \textbf{19.7} & 41.4 & \textbf{53.9} & 85.1 & \multicolumn{1}{c|}{\textbf{200.1}} & 8.7$\times$10$^4$ & \textbf{26.9} & \textbf{50.6} & 60.7 & \textbf{91.4} & \multicolumn{1}{c|}{229.6} & 1.1$\times$10$^4$ & 3.1 & \textbf{10.8} & \textbf{17.3} & \textbf{57.0} & \textbf{88.2} \\
\multicolumn{1}{c|}{$2\times2$} & 5.9$\times$10$^3$ & 19.3 & \textbf{41.5} & 53.8 & \textbf{85.3} & \multicolumn{1}{c|}{200.0} & 2.1$\times$10$^4$ & 26.7 & \textbf{50.6} & \textbf{61.4} & \textbf{91.4} & \multicolumn{1}{c|}{\textbf{230.1}} & 3.1$\times$10$^3$ & \textbf{3.3} & 10.6 & 17.0 & 55.2 & 86.0 \\
\multicolumn{1}{c|}{$3\times3$} & 2.7$\times$10$^3$ & 18.7 & 40.6 & 52.7 & 84.4 & \multicolumn{1}{c|}{196.4} & 9.8$\times$10$^3$ & 24.2 & 47.2 & 58.3 & 89.9 & \multicolumn{1}{c|}{219.6} & 1.4$\times$10$^3$ & 2.6 & 9.0 & 15.2 & 51.3 & 78.1 \\
\multicolumn{1}{c|}{$4\times4$} & 1.6$\times$10$^3$ & 16.8 & 38.1 & 50.0 & 83.4 & \multicolumn{1}{c|}{188.2} & 5.6$\times$10$^3$ & 21.6 & 43.3 & 54.3 & 88.7 & \multicolumn{1}{c|}{208.0} & 9.3$\times$10$^2$ & 2.3 & 7.1 & 11.9 & 44.3 & 65.6 \\
\multicolumn{1}{c|}{$5\times5$} & 1.1$\times$10$^3$ & 15.3 & 35.6 & 47.4 & 82.0 & \multicolumn{1}{c|}{180.3} & 3.6$\times$10$^3$ & 19.0 & 39.4 & 49.7 & 86.0 & \multicolumn{1}{c|}{194.1} & 7.0$\times$10$^2$ & 2.0 & 6.5 & 10.6 & 39.7 & 58.9 \\
\multicolumn{1}{c|}{$6\times6$} & 8.4$\times$10$^2$ & 13.6 & 33.0 & 44.9 & 80.0 & \multicolumn{1}{c|}{171.5} & 2.6$\times$10$^3$ & 15.9 & 35.0 & 45.2 & 83.5 & \multicolumn{1}{c|}{179.5} & 4.4$\times$10$^2$ & 1.4 & 5.1 & 8.4 & 35.4 & 50.3 \\
\hline
\end{tabular}
}
\label{tab:fine_tuned_overall}
\end{table*}

\begin{table*}[t]
    \centering
    \caption{Hybrid QASIR performance. \ScoreUp{Red} and \ScoreDown{blue} values indicate that scores are superior and inferior to MS-SL (B32), respectively.}
    \scalebox{0.57}{
    \begin{tabular}{lcccccccccccccccccc}
    \hline
    \rowcolor[HTML]{EFEFEF}
    \multicolumn{1}{l|}{\cellcolor[HTML]{EFEFEF}} & \multicolumn{6}{c|}{\cellcolor[HTML]{EFEFEF}ActivityNet Captions} & \multicolumn{6}{c|}{\cellcolor[HTML]{EFEFEF}TVR} & \multicolumn{6}{c}{\cellcolor[HTML]{EFEFEF}Charades-STA} \\ \hline
    \multicolumn{1}{c|}{} & GFLOPs & R@1 & R@5 & R@10 & R@100 & \multicolumn{1}{c|}{sumR} & GFLOPs & R@1 & R@5 & R@10 & R@100 & \multicolumn{1}{c|}{sumR} & GFLOPs & R@1 & R@5 & R@10 & R@100 & sumR \\ \hline
    \rowcolor[HTML]{EFEFEF}
    \multicolumn{1}{c|}{MS-SL (B32)} & 5.4$\times$10$^2$ & 11.7 & 31.2 & 43.5 & 81.7 & \multicolumn{1}{c|}{168.1} & 2.0$\times$10$^3$ & 17.5 & 39.5 & 51.3 & 88.3 & \multicolumn{1}{c|}{196.6} & 2.8$\times$10$^2$ & 1.2 & 4.4 & 7.6 & 41.5 & 54.7 \\
    \multicolumn{1}{c|}{MS-SL (L14)} & 9.8$\times$10$^3$ & 14.8 & 36.6 & 50.3 & 84.8 & \multicolumn{1}{c|}{186.5} & 3.7$\times$10$^4$ & 22.0 & 46.2 & 58.0 & 91.1 & \multicolumn{1}{c|}{217.3} & 5.0$\times$10$^3$ & 1.7 & 7.1 & 12.0 & 54.4 & 75.2 \\
    \multicolumn{1}{c|}{MS-SL (L14-336)} & 2.3$\times$10$^4$ & 14.2 & 36.9 & 50.4 & 85.2 & \multicolumn{1}{c|}{186.7} & 8.7$\times$10$^4$ & 24.7 & 49.3 & 60.6 & 92.0 & \multicolumn{1}{c|}{226.6} & 1.1$\times$10$^4$ & 2.3 & 8.6 & 14.1 & 57.6 & 82.6 \\ \hline
    \multicolumn{19}{l}{\cellcolor[HTML]{EFEFEF}High: $3\times3$ QASIR-B32, Low: $2\times2$ QASIR-L14/QASIR-L14-336} \\
    \multicolumn{1}{c|}{QASIR-L14} & \ScoreUp{2.7$\times$10$^2$} & \ScoreUp{18.0} & \ScoreUp{39.0} & \ScoreUp{51.2} & \ScoreDown{81.4} & \multicolumn{1}{c|}{\ScoreUp{189.5}} & \ScoreUp{1.7$\times$10$^3$} & \ScoreUp{23.0} & \ScoreUp{45.3} & \ScoreUp{56.1} & \ScoreUp{88.3} & \multicolumn{1}{c|}{\ScoreUp{212.7}} & \ScoreDown{8.3$\times$10$^2$} & \ScoreUp{2.6} & \ScoreUp{9.0} & \ScoreUp{14.4} & \ScoreUp{44.9} & \ScoreUp{70.9} \\
    \multicolumn{1}{c|}{QASIR-L14-336} & \ScoreDown{5.5$\times$10$^2$} & \ScoreUp{19.1} & \ScoreUp{40.8} & \ScoreUp{52.8} & 81.7 & \multicolumn{1}{c|}{\ScoreUp{194.3}} & \ScoreDown{4.1$\times$10$^3$} & \ScoreUp{26.6} & \ScoreUp{50.4} & \ScoreUp{61.1} & \ScoreUp{90.6} & \multicolumn{1}{c|}{\ScoreUp{228.6}} & \ScoreDown{1.9$\times$10$^3$} & \ScoreUp{3.1} & \ScoreUp{9.9} & \ScoreUp{15.8} & \ScoreUp{50.1} & \ScoreUp{78.9} \\
    \multicolumn{19}{l}{\cellcolor[HTML]{EFEFEF}High: $4\times4$ QASIR-B32, Low: $2\times2$ QASIR-L14/QASIR-L14-336} \\
    \multicolumn{1}{c|}{QASIR-L14} & \ScoreUp{2.4$\times$10$^2$} & \ScoreUp{17.8} & \ScoreUp{38.5} & \ScoreUp{50.6} & \ScoreDown{79.5} & \multicolumn{1}{c|}{\ScoreUp{186.4}} & \ScoreUp{1.7$\times$10$^3$} & \ScoreUp{22.9} & \ScoreUp{45.1} & \ScoreUp{55.8} & \ScoreDown{87.8} & \multicolumn{1}{c|}{\ScoreUp{211.6}} & \ScoreDown{8.2$\times$10$^2$} & \ScoreUp{2.5} & \ScoreUp{8.6} & \ScoreUp{13.8} & \ScoreUp{43.8} & \ScoreUp{68.8} \\
    \multicolumn{1}{c|}{QASIR-L14-336} & \ScoreUp{5.2$\times$10$^2$} & \ScoreUp{18.8} & \ScoreUp{40.3} & \ScoreUp{52.0} & \ScoreDown{79.8} & \multicolumn{1}{c|}{\ScoreUp{190.1}} & \ScoreDown{4.0$\times$10$^3$} & \ScoreUp{26.4} & \ScoreUp{50.0} & \ScoreUp{60.7} & \ScoreUp{90.0} & \multicolumn{1}{c|}{\ScoreUp{227.1}} & \ScoreDown{1.8$\times$10$^3$} & \ScoreUp{2.8} & \ScoreUp{9.4} & \ScoreUp{15.1} & \ScoreUp{48.5} & \ScoreUp{75.8} \\
    \multicolumn{19}{l}{\cellcolor[HTML]{EFEFEF}High: $5\times5$ QASIR-B32, Low: $2\times2$ QASIR-L14/QASIR-L14-336} \\
    \multicolumn{1}{c|}{QASIR-L14} & \ScoreUp{2.3$\times$10$^2$} & \ScoreUp{17.1} & \ScoreUp{37.0} & \ScoreUp{48.6} & \ScoreDown{75.9} & \multicolumn{1}{c|}{\ScoreUp{178.5}} & \ScoreUp{1.7$\times$10$^3$} & \ScoreUp{22.7} & \ScoreUp{44.5} & \ScoreUp{55.1} & \ScoreDown{85.9} & \multicolumn{1}{c|}{\ScoreUp{208.2}} & \ScoreDown{8.1$\times$10$^2$} & \ScoreUp{2.5} & \ScoreUp{8.4} & \ScoreUp{13.3} & \ScoreUp{41.7} & \ScoreUp{65.9} \\
    \multicolumn{1}{c|}{QASIR-L14-336} & \ScoreUp{5.1$\times$10$^2$} & \ScoreUp{18.1} & \ScoreUp{38.7} & \ScoreUp{50.0} & \ScoreDown{76.0} & \multicolumn{1}{c|}{\ScoreUp{182.7}} &\ScoreDown{4.0$\times$10$^3$} & \ScoreUp{23.0} & \ScoreUp{49.1} & \ScoreUp{59.5} & \ScoreDown{87.9} & \multicolumn{1}{c|}{\ScoreUp{222.5}} & \ScoreDown{1.8$\times$10$^3$} & \ScoreUp{2.8} & \ScoreUp{9.2} & \ScoreUp{14.6} & \ScoreUp{46.6} & \ScoreUp{73.2} \\
    \multicolumn{19}{l}{\cellcolor[HTML]{EFEFEF}High: $6\times6$ QASIR-B32, Low: $2\times2$ QASIR-L14/QASIR-L14-336} \\
    \multicolumn{1}{c|}{QASIR-L14} & \ScoreUp{2.2$\times$10$^2$} & \ScoreUp{16.5} & \ScoreUp{35.5} & \ScoreUp{46.4} & \ScoreDown{71.7} & \multicolumn{1}{c|}{\ScoreUp{170.2}} & \ScoreUp{1.7$\times$10$^3$} & \ScoreUp{22.3} & \ScoreUp{44.0} & \ScoreUp{54.5} & \ScoreDown{84.9} & \multicolumn{1}{c|}{\ScoreUp{205.7}} & \ScoreDown{8.1$\times$10$^2$} & \ScoreUp{2.3} & \ScoreUp{7.6} & \ScoreUp{12.2} & \ScoreDown{40.3} & \ScoreUp{62.5} \\
    \multicolumn{1}{c|}{QASIR-L14-336} & \ScoreUp{5.0$\times$10$^2$} & \ScoreUp{17.4} & \ScoreUp{37.1} & \ScoreUp{47.6} & \ScoreDown{71.8} & \multicolumn{1}{c|}{\ScoreUp{173.9}} & \ScoreDown{4.0$\times$10$^3$} & \ScoreUp{25.6} & \ScoreUp{48.5} & \ScoreUp{58.7} & \ScoreDown{86.8} & \multicolumn{1}{c|}{\ScoreUp{219.6}} & \ScoreDown{1.8$\times$10$^3$} & \ScoreUp{2.6} & \ScoreUp{8.7} & \ScoreUp{13.8} & \ScoreUp{44.9} & \ScoreUp{70.0} \\ \hline
    \end{tabular}
    }
\label{tab:hybrid_performance}
\end{table*}
\label{subsec:fine_tune_performance}
\noindent
\textbf{Fine-tuning QASIR enables VLMs to learn super images effectively (\tabref{tab:fine_tuned_overall}).} The $2\times2$ models mostly achieve comparable performance to the $1\times1$ models and heavyweight SOTA methods while reducing the computation costs. In addition, the performance gap decreases as the model size increases. For instance, the $2\times2$ QASIR-L14-336 attains a sumR of 200.0/230.1/86.0, closely approaching the performance of the $1\times1$ configuration.
\\\textbf{Hybrid QASIR minimizes the performance drop of large VLMs while significantly reducing computational costs (\tabref{tab:hybrid_performance}).} For instance, for ActivityNet Captions, using $3\times3$ QASIR-B32 and $2\times2$ QASIR-L14 results in just a slight decrease of 5.4 sumR while reducing GFLOPs by 90\% compared to using $2\times2$ QASIR-L14.
Additionally, our hybrid approaches consistently outperform MS-SL (B32) in terms of sumR performance across all datasets. For instance, when comparing the hybrid model of $3\times3$ QASIR-B32 and $2\times2$ QASIR-L14 with MS-SL (B-32), we observe a significant increase of 21.4/16.1/16.2 sumR, respectively. However, we also acknowledge two drawbacks: R@100 scores are lower on ActivityNet Captions/TVR, and GFLOPs is higher on Charades-STA.
Both drawbacks stem from the limited performance of the high-efficiency models. By enhancing their performance, we could observe an increase in R@100, facilitating accurate video retrieval with hybrid models. Furthermore, it would enable us to reduce $R$ to achieve lower GFLOPs for Charades-STA.
Thus, enhancing the performance of the high-efficiency model is crucial for overall performance.
\\\textbf{Performance change when varying the number of re-ranking target videos $R$.}
\begin{figure}[t]
  \centering
  \includegraphics[width=0.9\linewidth]{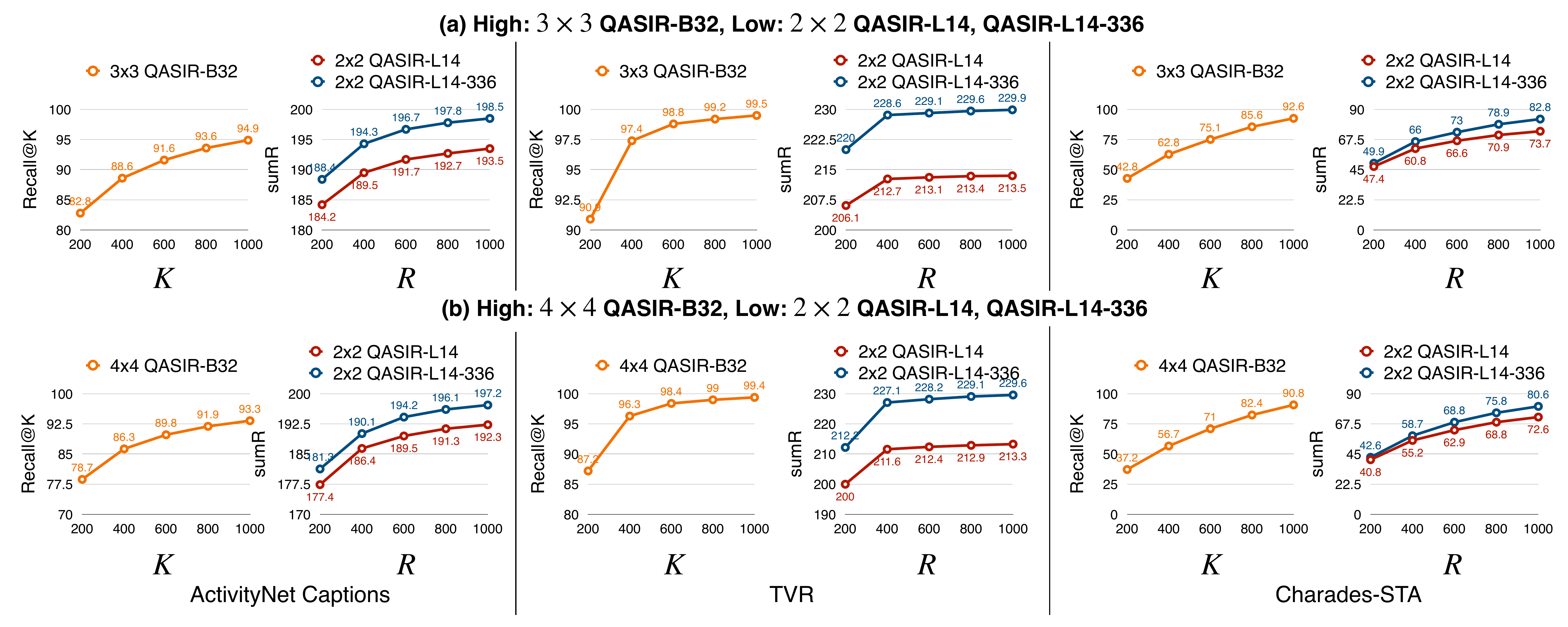}
  \caption{Recall@$K$ of the high-effcient model and sumR change when varying $K$ and $R$. (a) and (b) present the combination of $3\times3$/$2\times2$ and $4\times4$/$2\times2$ models, respectively.}
  \label{fig:performance_change_R}
\end{figure}
$R$ is an important trade-off parameter between the efficiency and retrieval performance.
\figref{fig:performance_change_R} demonstrates the Recall@$K$ of the high models and the sumR change of the hybrid models when varying $K$ and $R$.
In terms of the high models, we find that they work to screen out irrelevant videos, but the optimal $R$ varies depending on the dataset. For ActivityNet Captions and TVR, $R=400$ is adequate as Recall@$400$ achieves successful retrieval in 85\% and 95\% of cases, respectively. In contrast, for Charades-STA, $R=800$ is necessary to attain an 85\% retrieval success rate.
We notice a consistent trend in the performance change of hybrid models. For ActivityNet Captions/TVR, increasing $R$ from 200 to 400/600 notably enhances retrieval performance, but pushing $R$ further results in diminishing returns. For Charades-STA, boosting $R$ from 200 to 800 is necessary for improved retrieval performance.
Considering the associated rise in computational costs, settling for $R=400/600$ and $R=800$ strikes an optimal balance between the performance and efficiency for ActivityNet Captions/TVR and Charades-STA, respectively.
\\\textbf{Ablation study.}
\tabref{tab:ablation_study} shows an ablation study on the fine-tuned $2\times2$ QASIR-L14-336 on the ActivityNet Captions dataset.
The results indicate that all components contribute to the retrieval performance. When comparing the vision adapter, text adapter, and temporal encoder, two insights emerge.
First, the temporal encoder yields the highest performance increase, but it also incurs the highest increase in computational costs.
Second, the text adapter strikes a commendable balance between enhanced performance and computational costs. It achieves comparable performance to the temporal encoder, with only a slight decrease of -0.6 in sumR, while minimizing additional computational costs.
\begin{table}[t]
    \centering
    \caption{Ablation study. $\Delta_{flp}$, $\Delta_{par}$ represent the difference of GFLOPs and parameters (M) from zero-shot model, respectively.}
    \scalebox{0.6}{
    \begin{tabular}{l|ccccccc}
    \hline
     & $\Delta_{flp}$ & $\Delta_{par}$ & R@1 & R@5 & R@10 & R@100 & sumR \\ \hline
    Vision adapter (VA) & 1.1$\times$10$^{-4}$ & 0.13 & 16.0 & 36.2 & 48.2 & 82.8 & 183.0 \\
    Text adapter (TA) & 7.6$\times$10$^{-6}$ & 0.13 & 18.1 & 39.6 & 52.2 & 85.0 & 194.9 \\
    Temporal encoder (T) & 2.2$\times$10$^{-3}$ & 3.15 & 18.4 & 40.1 & 52.2 & 84.7 & 195.4 \\
    VA+TA & 1.2$\times$10$^{-4}$ & 0.26 & 19.1 & 41.2 & 53.4 & 84.8 & 197.6 \\
    VA+TA+T & 2.3$\times$10$^{-3}$ & 3.41 & \textbf{19.3} & \textbf{41.5} & \textbf{53.8} & \textbf{85.3} & \textbf{200.0} \\ \hline
    \end{tabular}
    }
    \label{tab:ablation_study}
\end{table}

\subsection{Comparison to other sampling approaches}
\label{subsec:comparison_to_other_approaches}
\begin{table}[t]
\centering
\caption{Comparison of super images with other reduce-then-encode methods: sparse sampling \cite{lei2021cvpr}, MGSampler \cite{Zhi_2021_ICCV}, and PMISampler \cite{Xian_2024_WACV}.}
\scalebox{0.54}{
\begin{tabular}{ccccccccccccccccccc}
\hline
\rowcolor[HTML]{EFEFEF} 
\multicolumn{1}{c|}{\cellcolor[HTML]{EFEFEF}} & \multicolumn{6}{c|}{\cellcolor[HTML]{EFEFEF}ActivityNet Captions} & \multicolumn{6}{c|}{\cellcolor[HTML]{EFEFEF}TVR} & \multicolumn{6}{c}{\cellcolor[HTML]{EFEFEF}Charades-STA} \\ \hline
\multicolumn{1}{c|}{} & GFLOPs & R@1 & R@5 & R@10 & R@100 & \multicolumn{1}{c|}{sumR} & GFLOPs & R@1 & R@5 & R@10 & R@100 & \multicolumn{1}{c|}{sumR} & GFLOPs & R@1 & R@5 & R@10 & R@100 & sumR \\ \hline
\rowcolor[HTML]{EFEFEF}
\multicolumn{19}{l}{\cellcolor[HTML]{EFEFEF}2$\times$2 setting (average \#Frames: 15.5/57.7/8.1 on ActivityNet Captions/TVR/Charades-STA)} \\
\multicolumn{1}{c|}{Sparse sampling} & & 18.0 & 39.4 & 51.6 & 83.7 & \multicolumn{1}{c|}{192.7} & & \textbf{27.1} & 49.6 & 60.1 & 91.3 & \multicolumn{1}{c|}{228.2} & & \textbf{3.3} & \textbf{11.0} & 16.2 & 54.0 & 84.5 \\
\multicolumn{1}{c|}{MGSampler} & & 18.1 & 39.5 & 51.6 & 83.5 & \multicolumn{1}{c|}{192.7} & & 26.7 & 49.9 & 60.4 & 90.9 & \multicolumn{1}{c|}{227.9} & & 3.0 & 9.7 & 15.2 & 54.5 & 82.3 \\
\multicolumn{1}{c|}{PMISampler} & & 17.9 & 39.3 & 51.5 & 83.7 & \multicolumn{1}{c|}{192.6} & & 26.9 & \textbf{50.6} & 60.7 & 91.1 & \multicolumn{1}{c|}{229.4} & & 3.0 & 10.1 & 16.0 & 53.3 & 82.4 \\
\multicolumn{1}{c|}{Super images} & \multirow{-4}{*}{5.9$\times$10$^3$} & \textbf{19.3} & \textbf{41.5} & \textbf{53.8} & \textbf{85.3} & \multicolumn{1}{c|}{\textbf{200.0}} & \multirow{-4}{*}{2.1$\times$10$^4$} & 26.7 & \textbf{50.6} & \textbf{61.4} & \textbf{91.4} & \multicolumn{1}{c|}{\textbf{230.1}} & \multirow{-4}{*}{3.1$\times$10$^3$} & \textbf{3.3} & 10.6 & \textbf{17.0} & \textbf{55.2} & \textbf{86.0} \\
\rowcolor[HTML]{EFEFEF}
\multicolumn{19}{l}{\cellcolor[HTML]{EFEFEF}3$\times$3 setting (average \#Frames: 7.1/23.9/3.9 on ActivityNet Captions/TVR/Charades-STA)} \\
\multicolumn{1}{c|}{Sparse sampling} & & 15.8 & 35.4 & 47.2 & 80.1 & \multicolumn{1}{c|}{178.5} & & \textbf{25.3} & \textbf{47.4} & 57.3 & 89.4 & \multicolumn{1}{c|}{219.4} & & 2.2 & 7.8 & 12.7 & 45.9 & 68.6 \\
\multicolumn{1}{c|}{MGSampler} & & 15.7 & 35.7 & 47.3 & 80.4 & \multicolumn{1}{c|}{179.2} & & 23.7 & 45.7 & 56.2 & 88.1 & \multicolumn{1}{c|}{213.7} & & 2.3 & 7.7 & 12.5 & 46.9 & 69.4 \\
\multicolumn{1}{c|}{PMISampler} & & 15.9 & 35.7 & 47.5 & 80.7 & \multicolumn{1}{c|}{179.8} & & 23.8 & 45.2 & 55.5 & 88.0 & \multicolumn{1}{c|}{212.6} & & 2.4 & 7.6 & 12.8 & 46.7 & 69.6 \\
\multicolumn{1}{c|}{Super images} & \multirow{-4}{*}{2.7$\times$10$^3$} & \textbf{18.7} & \textbf{40.6} & \textbf{52.7} & \textbf{84.4} & \multicolumn{1}{c|}{\textbf{196.4}} & \multirow{-4}{*}{9.8$\times$10$^3$} & 24.2 & 47.2 & \textbf{58.3} & \textbf{89.9} & \multicolumn{1}{c|}{\textbf{219.6}} & \multirow{-4}{*}{1.4$\times$10$^3$} & \textbf{2.6} & \textbf{9.0} & \textbf{15.2} & \textbf{51.3} & \textbf{78.1} \\
\rowcolor[HTML]{EFEFEF}
\multicolumn{19}{l}{\cellcolor[HTML]{EFEFEF}4$\times$4 setting (average \#Frames: 4.2/14.8/2.4 on ActivityNet Captions/TVR/Charades-STA)} \\
\multicolumn{1}{c|}{Sparse sampling} & & 13.2 & 30.9 & 41.8 & 74.8 & \multicolumn{1}{c|}{160.7} & & \textbf{22.8} & 43.2 & 53.5 & 86.7 & \multicolumn{1}{c|}{206.1} & & 1.8 & 6.3 & 10.9 & 40.5 & 59.6 \\
\multicolumn{1}{c|}{MGSampler} & & 10.4 & 31.9 & 43.7 & 77.1 & \multicolumn{1}{c|}{166.3} & & 21.8 & 42.7 & 52.6 & 85.8 & \multicolumn{1}{c|}{202.8} & & 1.8 & 6.9 & 11.3 & 39.1 & 59.1 \\
\multicolumn{1}{c|}{PMISampler} & & 13.7 & 32.0 & 43.3 & 77.0 & \multicolumn{1}{c|}{166.0} & & 22.4 & 43.1 & 53.3 & 86.7 & \multicolumn{1}{c|}{205.6} & & 1.8 & 6.6 & 10.8 & 39.4 & 58.6 \\
\multicolumn{1}{c|}{Super images} & \multirow{-4}{*}{1.6$\times$10$^3$} & \textbf{16.8} & \textbf{38.1} & \textbf{50.0} & \textbf{83.4} & \multicolumn{1}{c|}{\textbf{188.2}} & \multirow{-4}{*}{5.6$\times$10$^3$} & 21.6 & \textbf{43.3} & \textbf{54.3} & \textbf{88.7} & \multicolumn{1}{c|}{\textbf{208.0}} & \multirow{-4}{*}{9.3$\times$10$^2$} & \textbf{2.3} & \textbf{7.1} & \textbf{11.9} & \textbf{44.3} & \textbf{65.6} \\
\rowcolor[HTML]{EFEFEF} 
\multicolumn{19}{l}{\cellcolor[HTML]{EFEFEF}5$\times$5 setting (average \#Frames: 2.9/9.6/1.8 on ActivityNet Captions/TVR/Charades-STA)} \\
\multicolumn{1}{c|}{Sparse sampling} & & 11.3 & 26.7 & 36.3 & 66.4 & \multicolumn{1}{c|}{140.7} & & \textbf{19.4} & 38.7 & 48.7 & 83.1 & \multicolumn{1}{c|}{189.9} & & 1.6 & \textbf{6.6} & 10.0 & 35.1 & 53.2 \\
\multicolumn{1}{c|}{MGSampler} & & 12.2 & 28.7 & 38.8 & 70.7 & \multicolumn{1}{c|}{150.3} & & 19.1 & 38.6 & 48.7 & 83.0 & \multicolumn{1}{c|}{189.4} & & 1.9 & 6.5 & 10.0 & 34.3 & 52.6 \\
\multicolumn{1}{c|}{PMISampler} & & 12.0 & 28.6 & 39.0 & 70.6 & \multicolumn{1}{c|}{150.3} & & 19.3 & 38.2 & 48.0 & 83.0 & \multicolumn{1}{c|}{188.5} & & 1.8 & 6.3 & 10.2 & 34.6 & 52.8 \\
\multicolumn{1}{c|}{Super images} & \multirow{-4}{*}{1.1$\times$10$^3$} & \textbf{15.3} & \textbf{35.6} & \textbf{47.4} & \textbf{82.0} & \multicolumn{1}{c|}{\textbf{180.3}} & \multirow{-4}{*}{3.6$\times$10$^3$} & 19.0 & \textbf{39.4} & \textbf{49.7} & \textbf{86.0} & \multicolumn{1}{c|}{\textbf{194.1}} & \multirow{-4}{*}{7.0$\times$10$^2$} & \textbf{2.0} & 6.5 & \textbf{10.6} & \textbf{39.7} & \textbf{58.9} \\
\rowcolor[HTML]{EFEFEF} 
\multicolumn{19}{l}{\cellcolor[HTML]{EFEFEF}6$\times$6 setting (average \#Frames: 2.1/6.8/1.1 on ActivityNet Captions/TVR/Charades-STA)} \\
\multicolumn{1}{c|}{Sparse sampling} & & 9.6 & 22.7 & 30.7 & 57.1 & \multicolumn{1}{c|}{120.2} & & 16.2 & 33.6 & 42.8 & 79.6 & \multicolumn{1}{c|}{172.2} & & 1.8 & 5.9 & 10.1 & \textbf{34.7} & 52.6 \\
\multicolumn{1}{c|}{MGSampler} & & 10.8 & 25.4 & 34.7 & 63.2 & \multicolumn{1}{c|}{134.1} & & \textbf{17.1} & 33.7 & 42.8 & 78.6 & \multicolumn{1}{c|}{172.3} & & 1.8 & \textbf{6.6} & \textbf{10.2} & 33.9 & 52.5 \\
\multicolumn{1}{c|}{PMISampler} & & 10.3 & 25.1 & 34.2 & 62.9 & \multicolumn{1}{c|}{132.6} & & 16.7 & 33.9 & 42.8 & 78.4 & \multicolumn{1}{c|}{171.8} & & \textbf{1.9} & 6.3 & 9.9 & \textbf{34.7} & \textbf{52.9} \\
\multicolumn{1}{c|}{Super images} & \multirow{-4}{*}{8.4$\times$10$^2$} & \textbf{13.6} & \textbf{33.0} & \textbf{44.9} & \textbf{80.0} & \multicolumn{1}{c|}{\textbf{171.5}} & \multirow{-4}{*}{2.6$\times$10$^3$} & 15.9 & \textbf{35.0} & \textbf{45.2} & \textbf{83.5} & \multicolumn{1}{c|}{\textbf{179.5}} & \multirow{-4}{*}{4.4$\times$10$^2$} & 1.4 & 5.1 & 8.4 & 35.4 & 50.3 \\ \hline
\end{tabular}
}
\label{fig:performance_change_other_sampling}
\end{table}

To assess whether super images offer optimal efficiency in PRVR, we compare super images with other reduce-then-encode sampling methods: sparse sampling \cite{lei2021cvpr}, MGSampler \cite{Zhi_2021_ICCV}, and PMISampler \cite{Xian_2024_WACV}.
We employ the QASIR-L14-336 backbone and apply these methods to $1\times1$ QASIR-L14-336.
To ensure a fair evaluation of computational efficiency, we count the number of super images $L$ for each grid size in every video. We then train the $1\times1$ models by sampling $L$ frames per video using each sampling algorithm. Therefore, the number of encoding frames and GFLOPs remain consistent across the methods.
\tabref{fig:performance_change_other_sampling} presents the results, indicating that super images perform better than other approaches and mitigate the performance degradation in general, particularly evident when reducing the number of encoding frames to the $N=5,6$ settings.
The inferior performance for Charades-STA at $N=6$ may stem from the limited visibility of objects (see Appendix G).

\subsection{Moment-to-video performance}
\label{subsec:moment_video_performance}
\begin{figure}[t]
  \centering
  \includegraphics[width=0.8\linewidth]{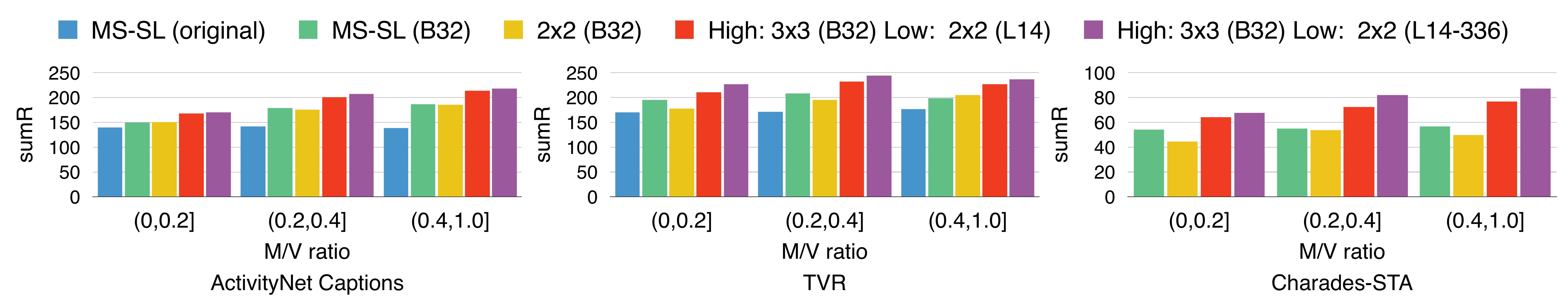}
  \caption{Moment-to-video performance on benchmark datasets. Note that we cannot evaluate M/V performance on MS-SL (original) on Charades-STA because original feature files and model weights are currently unavailable.}
  \label{fig:moment_to_video}
\end{figure}

As with previous PRVR studies \cite{dong2022acmmm,dong2023iccv}, we report grouped sumR using the moment-to-video ratio (M/V), which is computed as a moment's length ratio in the entire video duration.
A smaller M/V indicates that the videos have less relevant moments with irrelevant content, thereby retrieving them from the corresponding queries is more challenging.
As with \cite{dong2023iccv}, we compute the sumR scores for three M/V settings, where the moments are short ($r \in (0,0,2]$), middle ($r \in (0.2,0.4]$), and long ($r \in (0.4,1.0]$).
\figref{fig:moment_to_video} presents the M/V performance, indicating that the hybrid QASIR outperforms the baselines for all of the settings.
When comparing QASIR across different M/V ratios, we observe improved performance as the ratio increases. This suggests that super images excel at retrieving videos containing middle/long moments, rather than short ones.

\subsection{Qualitative results}

\begin{figure*}[t]
  \centering
  \includegraphics[width=0.8\linewidth]{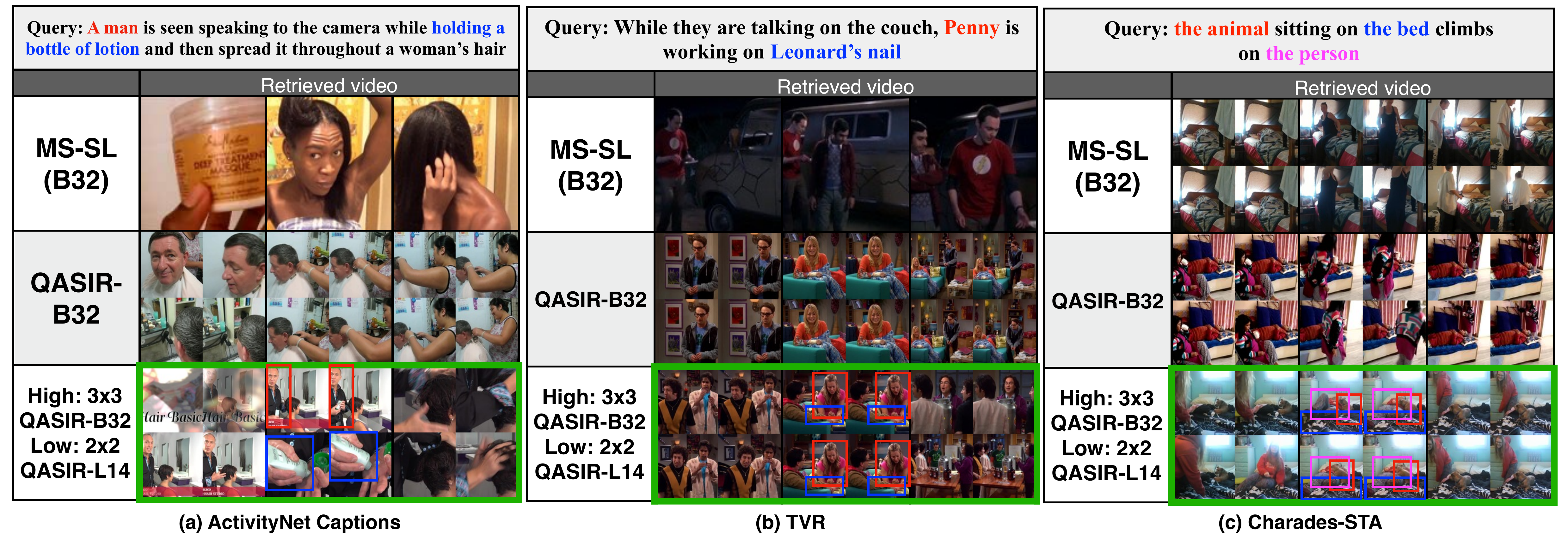}
  \caption{Examples of videos retrieved from different models. Colored rectangles represent objects corresponding to words in input query.}
  \label{fig:retrieved_results}
\end{figure*}

\figref{fig:retrieved_results} shows examples of videos retrieved from different three models.
Although MS-SL (B32) and QASIR-B32 accurately identify objects in videos, they fall short of capturing their relationships. For instance, in (a), they successfully identify ``a man'' and ``a woman's hair'' but fail to grasp the relationship between the man holding a bottle of lotion and applying it to the woman's hair. With larger VLM capabilities, the hybrid model can capture such concepts and retrieve videos that match the corresponding query. A discussion on failure cases is written in Appendix H.

\subsection{Are super images effective for other video-language tasks?}
\begin{table}[t]
\centering
\caption{Performance on T2VR and video captioning. \textbf{Bold} represents the higher of sparse sampling and super images. B, M, R, C represents BLEU \cite{papineni2002acl}, METEOR \cite{banerjee2005acl}, ROUGE-L \cite{lin2004acl}, and CIDEr-D \cite{vedantam2015cvpr}, respectively.}
\scalebox{0.5}{
\begin{tabular}{lcccccccc|lcccccccccc}
\hline
\rowcolor[HTML]{EFEFEF} 
\multicolumn{9}{c|}{\cellcolor[HTML]{EFEFEF}Traditional T2VR} & \multicolumn{11}{c}{\cellcolor[HTML]{EFEFEF}Video captioning} \\ \hline
\rowcolor[HTML]{EFEFEF} 
\multicolumn{1}{c}{\cellcolor[HTML]{EFEFEF}} & \multicolumn{4}{c}{\cellcolor[HTML]{EFEFEF}MSRVTT} & \multicolumn{4}{c|}{\cellcolor[HTML]{EFEFEF}ActivityNet Captions} &  & \multicolumn{5}{c}{\cellcolor[HTML]{EFEFEF}MSRVTT} & \multicolumn{5}{c}{\cellcolor[HTML]{EFEFEF}MSVD} \\ \hline
\multicolumn{1}{c}{} & \#Frames & R@1 & R@5 & R@10 & \#Frames & R@1 & R@5 & R@50 &  & \#Frames & B4 & M & R & C & \#Frames & B4 & M & R & C \\ \hline
CLIP4Clip \cite{luo2021arxiv} & 12 & 43.1 & 70.4 & 80.8 & 64 & 40.5 & 72.4 & 98.1 & SwinBERT \cite{Lin_2022_CVPR} & 32 & 41.8 & 29.9 & 62.1 & 54.4 & 32 & 59.5 & 41.8 & 78.3 & 122.5 \\
+ sparse sampling & 3 & 34.5 & 63.2 & 73.1 & 16 & 37.0 & 68.1 & 96.7 & + sparse sampling & 8 & 28.2 & 23.9 & 55.1 & 26.8 & 8 & 46.8 & 33.1 & 71.3 & 75.5 \\
\rowcolor[HTML]{EFEFEF}
+ $2\times2$ super images & 3 & \textbf{37.9} & \textbf{65.8} & \textbf{77.2} & 16 & \textbf{38.3} & \textbf{69.5} & \textbf{96.8} & + $2\times2$ super images & 8 & \textbf{31.3} & \textbf{24.9} & \textbf{56.5} & \textbf{29.2} & 8 & \textbf{54.4} & \textbf{33.9} & \textbf{71.6} & \textbf{78.3} \\ \hline
\end{tabular}
}
\label{tab:other_tasks}
\end{table}
Finally, to address this question, we report the results for the traditional T2VR and video captioning. As the base models, we utilize CLIP4Clip-meanP (w/ CLIP-B/32) \cite{luo2021arxiv} and SwinBERT \cite{Lin_2022_CVPR} as the T2VR and video captioning.
Both of them achieve comparable performance to SOTA methods.
\tabref{tab:other_tasks} presents the results. Compared with sparse sampling, super images effectively mitigate performance degradation while reducing computational costs.
However, we acknowledge performance gaps compared with the base models. Our future work is to address them by delving into super images for versatile video-language tasks.

\section{Conclusion}
\label{sec:conclusion}

This paper proposed QASIR, which combines super images and VLMs for efficient and effective PRVR.
The main findings are three-fold. First, the zero-shot QASIR revealed that with a simple query-image attention trick, the VLMs generalize to super images. Second, the fine-tuning QASIR indicated that VLMs learn super images effectively and achieve comparable performance to $1\times1$ while reducing the computation costs. Finally, the hybrid QASIR showed that they strike a good balance between performance and computation costs.
\\\noindent
\textbf{Limitations.} This work has two limitations: (1) the performance bottleneck of the high-efficiency models (\secref{subsec:fine_tune_performance}) and (2) the reduced performance in retrieving videos with a small M/V ratio (\secref{subsec:moment_video_performance}).
To tackle (1), one possible approach is to implement KD between the high- and low-efficiency models.  As for (2), a potential solution is to extend QASIR to the spatio-temporal attentive models by leveraging patch-wise representations from Vision Transformer \cite{dosovitskiy2021an}.

\renewcommand{\thesection}{\Alph{section}}

\section*{Appendix}
\setcounter{section}{0}

\section{MLP architecture of the feature adapters}
\label{sec:adapter_detail_MLPs}

\begin{figure}[h]
  \centering
  \includegraphics[width=0.7\linewidth]{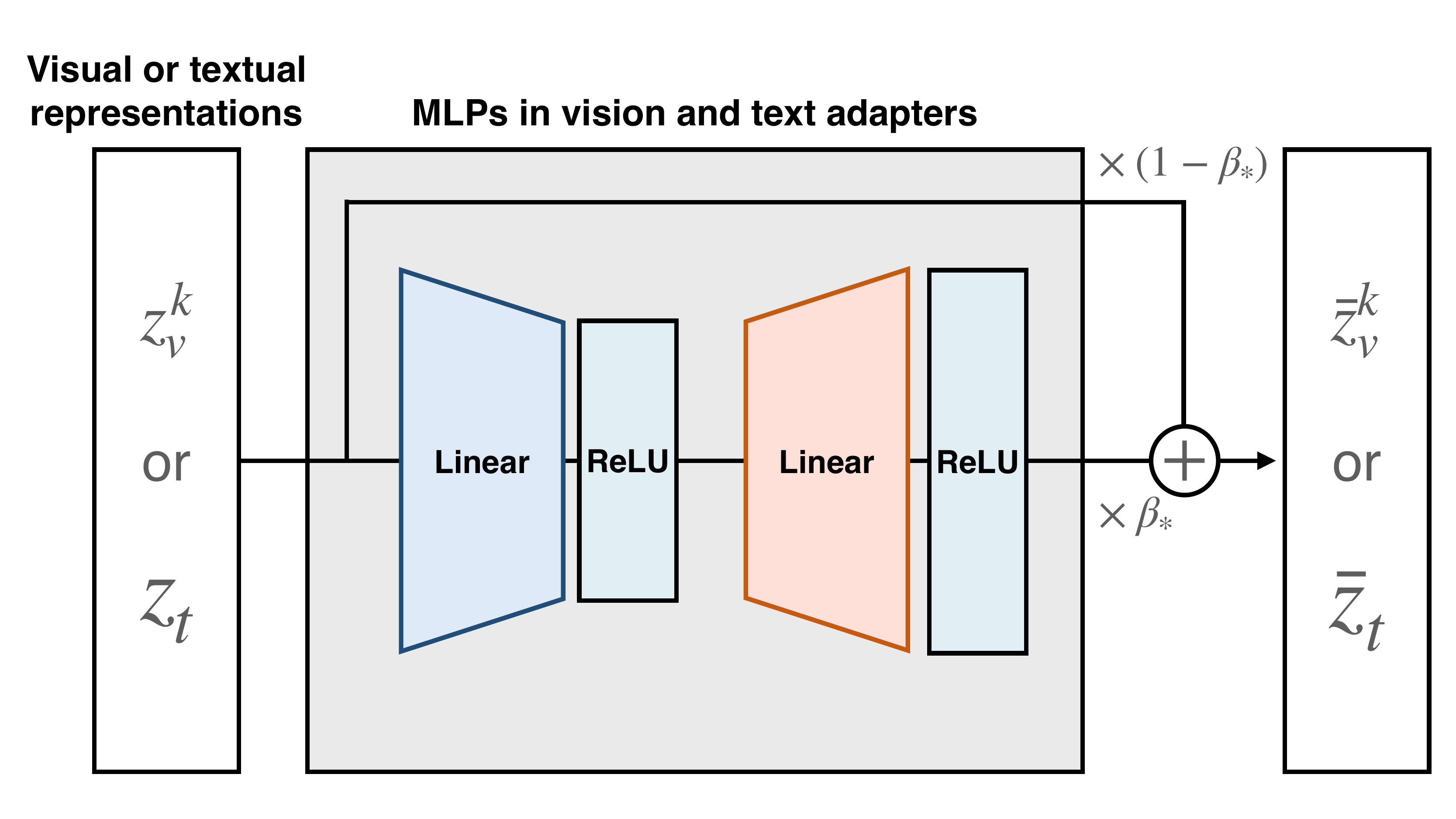}
  \caption{MLP architecture used in the vision and text adapter. They consist of two layers of linear with ReLU activation functions.}
  \label{fig:mlp_architecture}
\end{figure}

\figref{fig:mlp_architecture} shows the detailed architecture of MLPs in the vision adapter and text adapter.
As with \cite{gao2023ijcv}, their architecture is an auto encoder, consisting of two layers of linear with the ReLU activation function.
Given visual and textual representations $\Bdma{z}_v^k$ and $\Bdma{z}_t$, the adapters compute their intermediate representations $\Bdma{\bar{z}}_v^k$ and $\Bdma{\bar{z}}_t$ by interpolating input vectors and outputs of MLPs.

\section{Dataset statistics}

We describe the detailed statistics on the benchmark datasets.

\begin{itemize}
    \item \textbf{ActivityNet Captions} \cite{krishna2017iccv} collected YouTube videos and annotated them with event and sentence pairs. The average length of videos is 117.6 seconds, which is the longest among the datasets. Following \cite{dong2022acmmm,dong2023iccv}, 10,009 videos with 37,421 events are used for training and 4,917 videos with 17,505 events are used for testing.
    \item \textbf{TV show Retrieval (TVR)} \cite{lei2020eccv} contains 21.8K TV show videos with five sentences that describe a specific event in the video. The video duration is 76.2 seconds on average. As with \cite{dong2022acmmm,dong2023iccv}, 17,435 videos with 87,175 events are used for training and 2,179 videos with 10,895 events are used for testing.
    \item \textbf{Charades-STA} \cite{gao2017iccv} contains 6,670 videos with 16,128 sentences. The average video duration is 30.0 seconds. We follow the standard split as with \cite{dong2022acmmm,dong2023iccv}. Note that experiments on the Charades-STA are described in the supplementary materials due to space limitations.
\end{itemize}
Note that event timestamp annotations are not used for PRVR.

\section{Accurate definition of video-text GFLOPs}
The current PRVR methods \cite{dong2022acmmm,dong2023iccv}, including ours, contain three phases to compute video-text similarity: (1) visual and textual backbones extract features, (2) visual and textual encoders map them into a joint embedding space, and (3) matching module computes their similarity.
The video-text GFLOPs computes the total number of floating point operations of these three processes.

We formulate it using mathematical notations.
Let $\Bdma{y}$ be a query sentence and $\Bdma{X}=(\Bdma{x}_1, \ldots, \Bdma{x}_l, \ldots, \Bdma{x}_L)$ be a video containing $L$ frames where $\Bdma{x}_l$ is a $l$-th frame (image) in the test set $\mathcal{D}_{test}$.
Given $(\Bdma{X},\Bdma{y})$, visual and textual backbones $B_x,B_y$ convert them into intermediate representations $\Bdma{z}_x,\Bdma{z}_y$.
The visual and textual encoders $E_x,E_y$ further map them into a joint embedding space, yielding their embedded representations $\Bdma{\hat{z}}_x,\Bdma{\hat{z}}_y$. Finally, based on them, the matching module $M$ computes their similarity.
The video-text GFLOPs $G_{vt}$ computes the average of the total GFLOPs of all of these processes in video-text pairs in the test set, and is defined as:
\begin{align}
G_{vt}(\Bdma{X},\Bdma{Y}) &= \nonumber \\
\frac{1}{|\mathcal{D}_{test}|} & \sum_{(\Bdma{X},\Bdma{y}) \in \mathcal{D}_{test}} \biggl\{\underbrace{{\color{red} \sum_{l=1}^{L}G(B_x(\Bdma{x}_l))} + G(B_y(\Bdma{y})}_{\text{backbone GFLOPs}}) \nonumber \\
&+ \underbrace{G(E_x(\Bdma{z}_x)) + G(E_y(\Bdma{z}_y))}_{\text{encoder GFLOPs}} \nonumber \\
&+ \underbrace{G(M(\Bdma{\hat{z}}_x,\Bdma{\hat{z}}_y))}_{\text{matching module GFLOPs}}\biggr\},
\end{align}
where $|D_{test}|$ represents the number of total pairs of a video and query in the test set and $G(\cdot)$ computes the GFLOPs of each process.
Note that the red-term GFLOPs of image backbone encoding is much higher than others especially for long untrimmed videos because increasing frames $L$ yields increased GFLOPs. The super images address this problem by reducing the number of such image backbone encodings.

\section{Detailed settings on the conventional backbones}
On the visual side, I3D \cite{carreira2017cvpr} features extracted at 1 FPS are used for ActivityNet Captions and Charades-STA, and concatenated features of I3D and ResNet152 \cite{he2016cvpr} at 15 FPS are utilized for TVR.
For the textual side, they use RoBERTa \cite{liu2019roberta} for query feature extraction for both datasets.
For TVR, RoBERTa was additionally fine-tuned on the queries and subtitle sentences of TVR.
DL-DKD uses CLIP-B/32 only for the training phase to transfer CLIP's knowledge into the visual/textual features through a knowledge distillation framework.

\section{Implementation details}

We set 0.5 FPS for ActivityNet Captions and 1 FPS for Charades-STA to obtain frames from the videos. For TVR, we use the frames extracted at 3 FPS officially distributed by the TVR authors. The raw videos are not due to copyright issues.
For both vision and text adapters, we set the $\beta_v=\beta_t=0.2$ as with \cite{gao2023ijcv} and hidden size to be 192.
We fine-tune the model using the AdamW optimizer \cite{loshchilov2018iclr} with an initial learning rate of $0.0001$. The batch size is set to 64.

\begin{table}[t]
\centering
\caption{Zero-shot retrieval performance on each TV show. We use the $2\times2$ QASIR-L14-336 as our base model.}
\begin{tabular}{ccccccc}
\hline
\rowcolor[HTML]{EFEFEF} 
 & \multicolumn{6}{c}{\cellcolor[HTML]{EFEFEF}TVR} \\ \hline
TV show title & \#Videos & R@1 & R@5 & R@10 & R@100 & sumR \\ \hline
The Big Bang Theory & 2,155 & 22.6 & 42.6 & 53.1 & 84.4 & 202.7 \\
How I Met Your Mother & 745 & 24.8 & 43.4 & 54.6 & 87.2 & 210.1 \\
Friends & 2,800 & 19.5 & 36.8 & 46.4 & 81.0 & 183.8 \\
Grey's Anatomy & 520 & 15.9 & 32.1 & 40.2 & 78.1 & 166.3 \\
House & 2,310 & 11.9 & 24.3 & 31.6 & 66.0 & 133.9 \\
Castle & 2,365 & 12.4 & 26.6 & 33.8 & 68.7 & 141.5 \\ \hline
Total & 10,835 & 17.2 & 33.3 & 42.1 & 76.1 & 168.7 \\ \hline
\end{tabular}
\label{tab:tvshow_results}
\end{table}

\section{Performance gap analysis across ActivityNet Captions/Charades-STA and TVR}
\label{subsec:performance_gap_analysis}
\begin{figure}[t]
  \centering
  \includegraphics[width=0.6\linewidth]{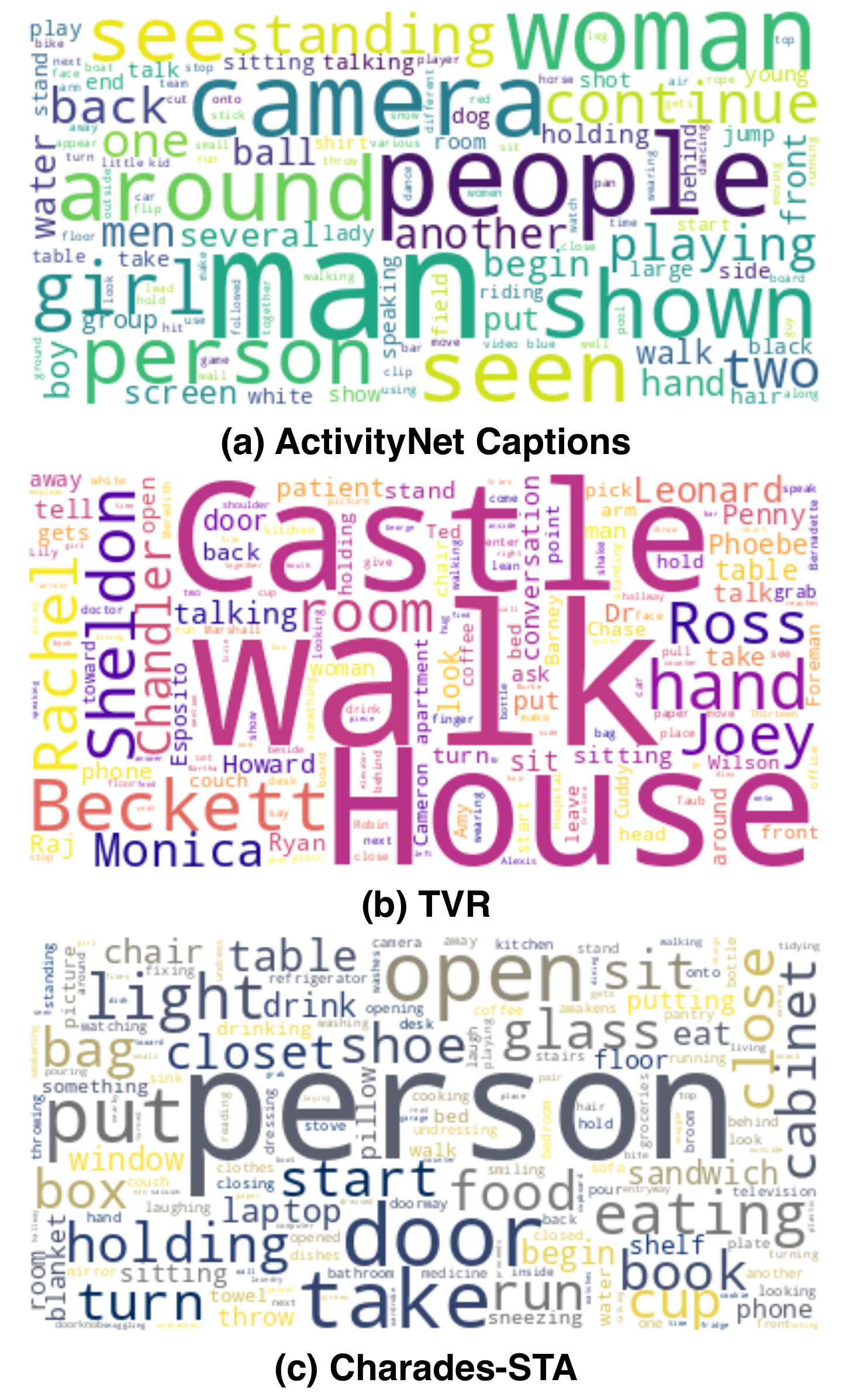}
  \caption{Word clouds for the benchmark datasets. The size of each word is proportional to the frequency in the dataset.}
  \label{fig:wordcloud}
\end{figure}
In Section 4.2 in the main paper, we assume that the performance difference across the datasets arises due to the noun gap.
To demonstrate this, we show word clouds\footnote{To generate word clouds, we use this library: \url{https://amueller.github.io/word_cloud/}} generated from query sentences in the test set on each dataset (\figref{fig:wordcloud}). As the word clouds suggest, general objects, such as ``man,'' ``woman,'' and ``person'' appear frequently in the ActivityNet Captions/Charades-STA \cite{krishna2017iccv,gao2017iccv}, while the character names, such as ``Beckett,'' ``Monica,'' and ``Sheldon'' appear frequently in TVR \cite{lei2020eccv}.
Therefore, if VLMs do not know the link between these names and their visual representations, they cannot retrieve the corresponding videos from input queries successfully. This is the reason for decreased zero-shot performance in TVR and increased fine-tuning performance by enabling models to understand their linking and outperform previous SOTA methods.
\\\textbf{Performance across TV shows.}
Then, a question arises, \textit{how much do VLMs know the TV shows that TVR covers?} If VLMs do not see them completely, VLMs perform poorly because of disjoint between characters and their visual representations. To investigate them, we aggregate the zero-shot performance on each TV show.
\tabref{tab:tvshow_results} shows the result, presenting a large performance gap across TV shows.
There is a 76.2 sumR gap between the highest (How I Met Your Mother) and lowest (House).
Therefore, we infer that VLMs possess knowledge of these TV shows to some extent, but their understanding is influenced by biases that stem from their training dataset.
Because the VLM backbone is OpenAI CLIP-L/14-336 \cite{radford2021icml} pre-trained on private data, further analysis of the training data to determine its inclusion of images from these TV shows is infeasible.

\section{Discussion on the inferior performance in Charades-STA at $N=6$}
\begin{figure}[t]
  \centering
  \includegraphics[width=\linewidth]{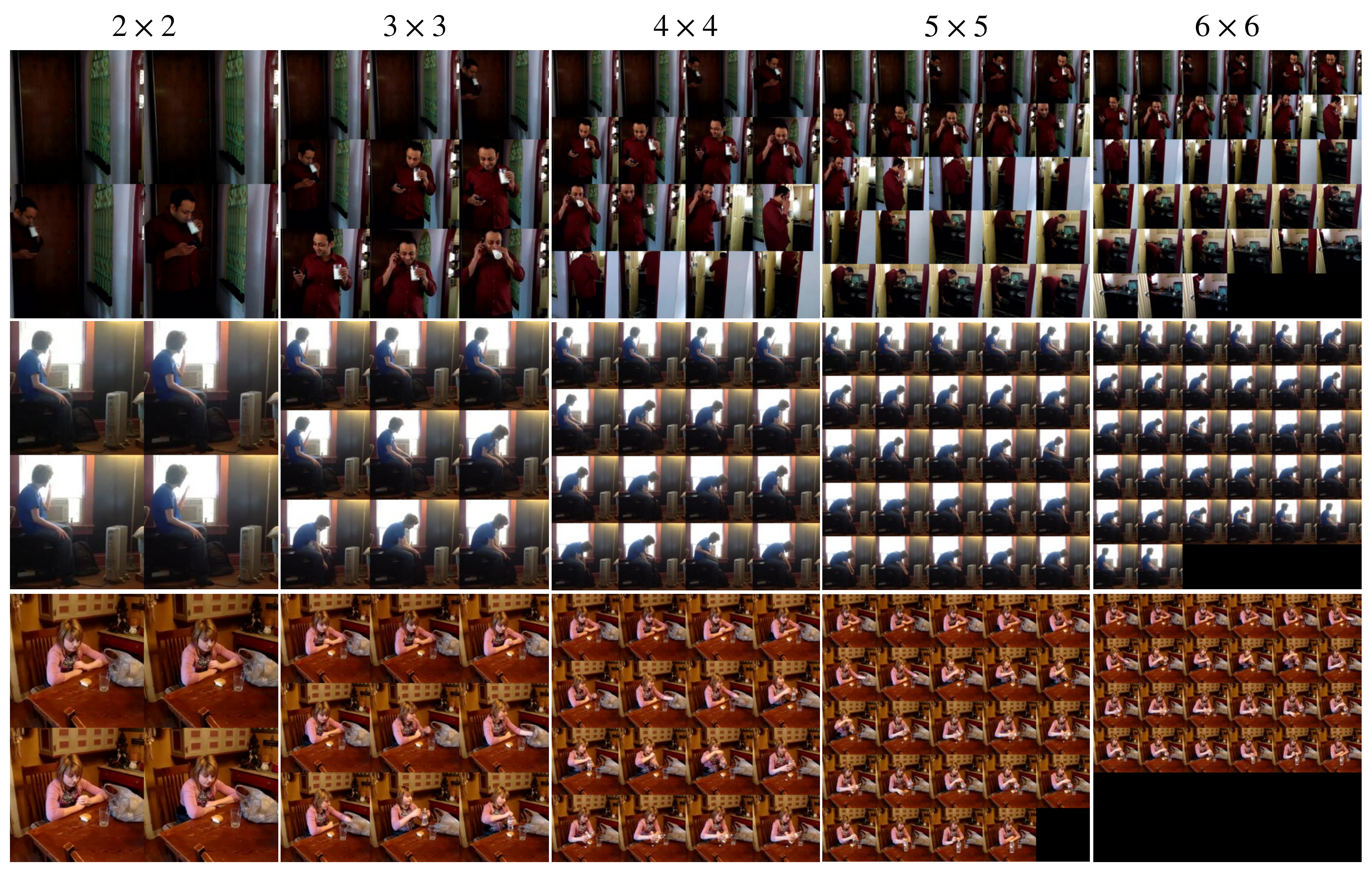}
  \caption{Super image examples in the Charades-STA dataset.}
  \label{fig:super_image_charades}
\end{figure}
As shown in Table 6 in the main paper, $6\times6$ super images perform worse than other sampling methods in Charades-STA. To investigate this, we randomly sample super images from the Charades-STA dataset by changing $N$ from 2 to 6. \figref{fig:super_image_charades} shows a comparison of super images in Charades-STA. When $N=2$, objects are easily recognizable (e.g., caps in the first row), but at $N=6$, they become small to identify. Therefore, we assume that this decreased visibility yields the difficulty of retrieving videos accurately.

\section{Failure retrieval cases}
\begin{figure}[t]
  \centering
  \includegraphics[width=\linewidth]{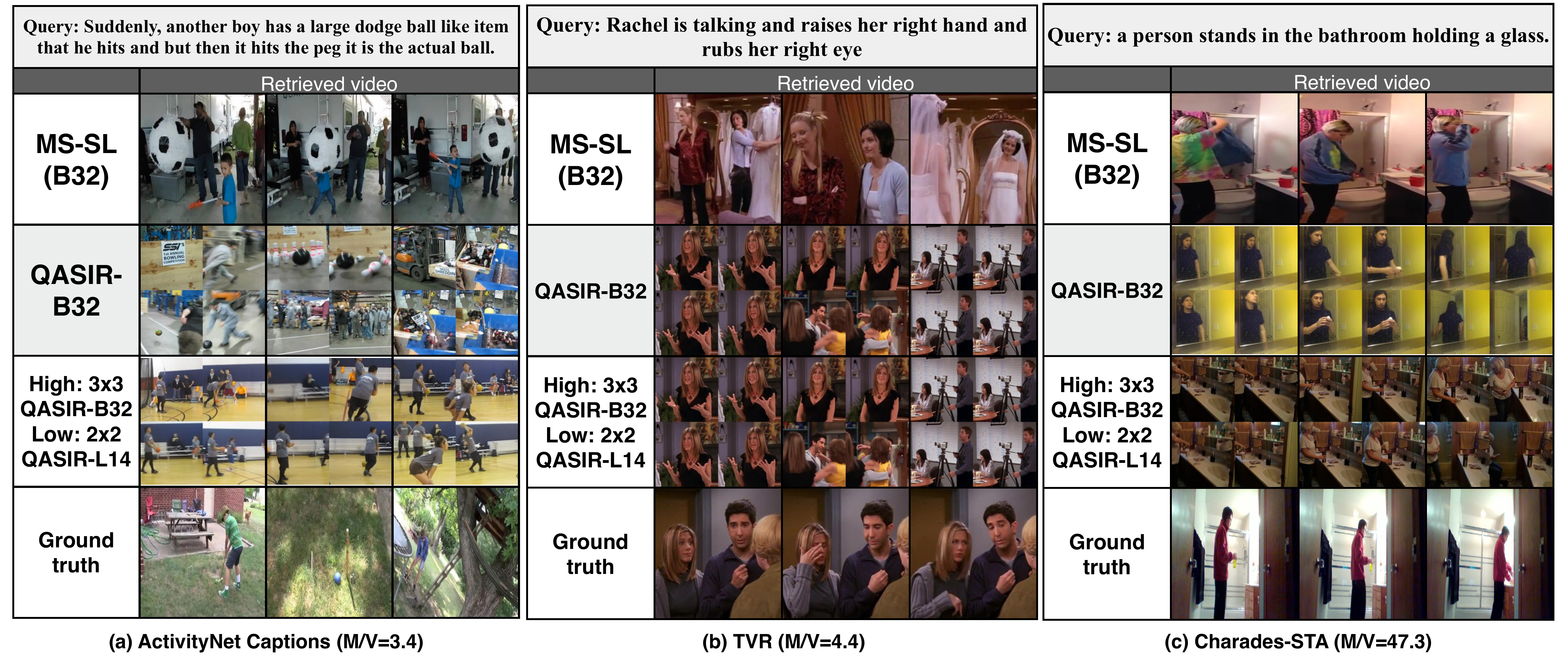}
  \caption{Failure cases on benchmark datasets.}
  \label{fig:failure_cases}
\end{figure}

\figref{fig:failure_cases} shows failure instances on benchmark datasets. It is evident that models struggle particularly in two cases: (1) when dealing with videos possessing a low M/V ratio (as elaborated in Section 4.5), and (2) when the model fails to grasp the complete essence of queries.
Examples (a) and (b) correspond to the former case due to their small M/V ratios.
Example (c) corresponds to the latter case. MS-SL, and the hybrid model capture ``a person,'' ``glass,'' and ``bathroom,'' but fail to capture ``holding.''
As mentioned in Section 5, one possible solution is to extend QASIR into spatio-temporal attention models to learn fine-grained correspondence between the query and super images.

\bibliographystyle{splncs04}
\bibliography{egbib}
\end{document}